\documentclass[sigconf]{acmart}

\AtBeginDocument{%
  }

\copyrightyear{2025}
\acmYear{2025}
\setcopyright{cc}
\setcctype{by-nc}
\acmConference[CIKM '25]{Proceedings of the 34th ACM International Conference on Information and Knowledge Management}{November 10--14, 2025}{Seoul, Republic of Korea}
\acmBooktitle{Proceedings of the 34th ACM International Conference on Information and Knowledge Management (CIKM '25), November 10--14, 2025, Seoul, Republic of Korea}\acmDOI{10.1145/3746252.3761345}
\acmISBN{979-8-4007-2040-6/2025/11}

\usepackage{titletoc}
\usepackage{soul}
\usepackage{hhline}
\usepackage{multirow}
\usepackage{graphicx}
\usepackage{pifont}
\usepackage{amsmath}
\usepackage{makecell}
\usepackage{multirow}
\usepackage{colortbl}
\usepackage{array}
\usepackage{rotating}
\usepackage{wrapfig}
\usepackage{enumitem}
\usepackage{float}
\usepackage{setspace}
\usepackage{placeins}
\usepackage{subcaption}
\usepackage[most]{tcolorbox}

\usepackage{tabularx}
\usepackage{multirow}
\usepackage{array}

\usepackage[dvipsnames]{xcolor}

\begin{document}


\title{FakeChain: Exposing Shallow Cues in Multi-Step Deepfake Detection}

\author{Minji Heo}
\affiliation{%
  \institution{Sungkyunkwan University}
  \department{Department of Artificial Intelligence}  
  \city{Suwon}
  \state{Gyeonggi-do}
  \country{South Korea}}
\email{minji.h0224@g.skku.edu}

\author{Simon S. Woo}
\authornote{Corresponding Author}
\affiliation{%
  \institution{Sungkyunkwan University}
  \department{Computer Science \& Engineering Department}
  \city{Suwon}
  \state{Gyeonggi-do}
  \country{South Korea}}
\email{swoo@g.skku.edu}

\renewcommand{\shortauthors}{Minji Heo and Simon S. Woo}
\begin{abstract}

Multi-step or hybrid deepfakes, created by sequentially applying different deepfake creation methods such as Face-Swapping, GAN-based generation, and Diffusion methods, can pose an emerging and unforseen technical challenge for detection models trained on single-step forgeries. While prior studies have mainly focused on detecting isolated single manipulation, little is known about the detection model behavior under such compositional, hybrid, and complex manipulation pipelines. In this work, we introduce \textbf{FakeChain}, a large-scale benchmark comprising 1-, 2-, and 3-Step forgeries synthesized using five state-of-the-art representative generators. Using this approach, we analyze detection performance and spectral properties across hybrid manipulation at different step, along with varying generator combinations and quality settings. Surprisingly, our findings reveal that detection performance highly depends on the final manipulation type, with F1-score dropping by up to \textbf{58.83\%} when it differs from training distribution. This clearly demonstrates that detectors rely on last-stage artifacts rather than cumulative manipulation traces, limiting generalization. Such findings highlight the need for detection models to explicitly consider manipulation history and sequences. Our results highlight the importance of benchmarks such as FakeChain, reflecting growing synthesis complexity and diversity in real-world scenarios. Our sample code is available here\footnote{\url{https://github.com/minjihh/FakeChain}}. 

\end{abstract}

\ccsdesc[500]{Computing methodologies~Image manipulation}
\ccsdesc[500]{Security and privacy~Social aspects of security and privacy}

\keywords{Deepfake detection; Multi-step manipulation; Multimedia forensics}


\maketitle

\begin{figure}[t]
  \centering
  \begin{subfigure}[b]{1\linewidth}
    \centering
    \includegraphics[trim={1pt 5pt 5pt 0pt},clip,width=1\linewidth]{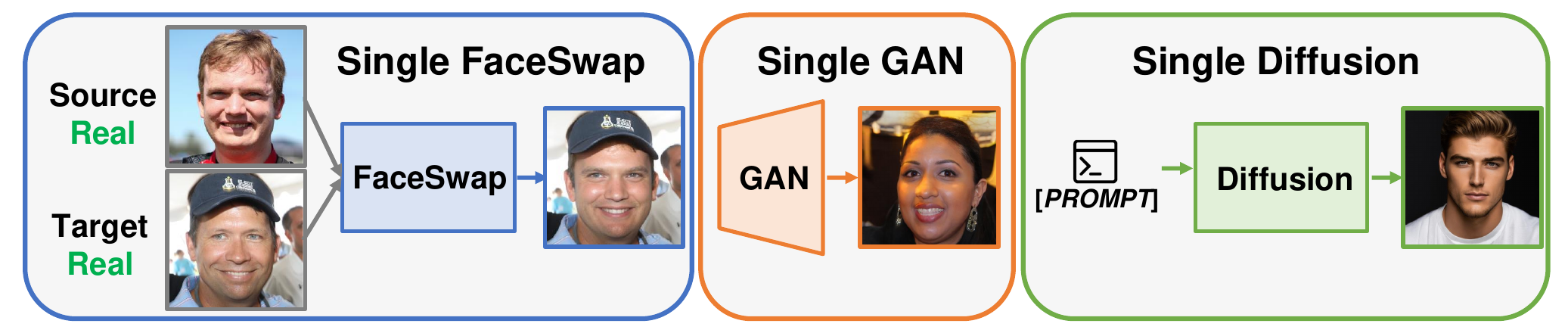}
    \caption{Traditional Fake Generation}
    \label{fig:traditional}
  \end{subfigure}

  \begin{subfigure}[b]{1\linewidth}
    \centering
    \includegraphics[trim={1pt 5pt 5pt 0pt},clip,width=1\linewidth]{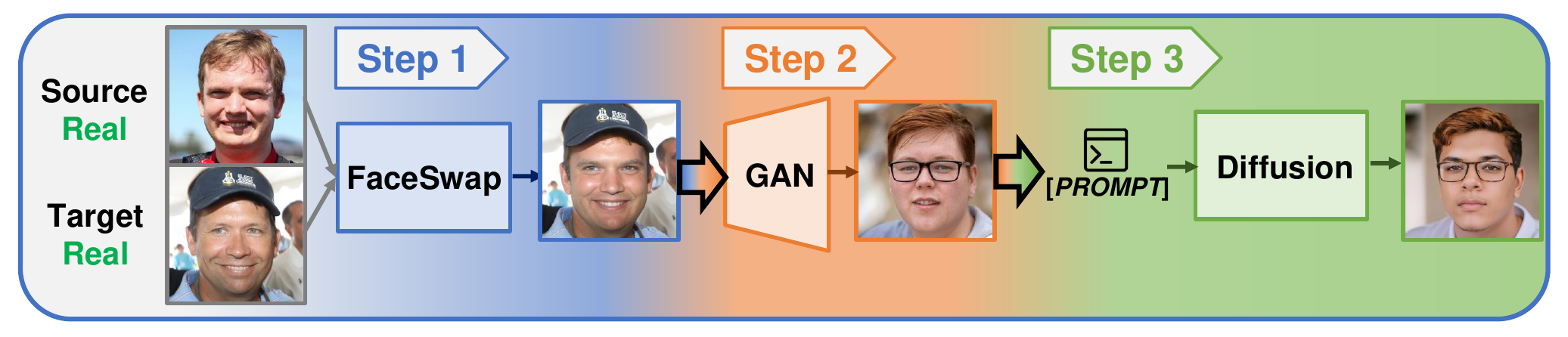}
    \caption{Proposed Multi-Step Manipulation}
    \label{fig:proposed}
  \end{subfigure}

  \caption{Comparison between Single-Step and Multi-Step Deepfake Generation \& Manipulation (FakeChain) Pipelines.}
  \label{fig:singlevsmulti}
\end{figure}

\section{Introduction}

The rapid evolution of generative models has transformed the landscape of synthetic media, significantly enhancing the realism, diversity, and accessibility of deepfakes. Key manipulation approaches include early Face-Swapping frameworks~\cite{faceswap2018}, GAN-based models~\cite{karras2019style}, and more recently, Diffusion models~\cite{ho2020ddpm,rombach2022ldm}, all of which have collectively pushed and significantly advanced the boundaries of synthetic realism. 

So far, the deepfake generation method has been applied once in the majority of detection research works
~\cite{1,2,5,7,9,18,20,26,30,33,35,39}. However, malicious users can apply deepfake creation methods sequentially for the same source input.  
For example, a face can be swapped using a GAN and subsequently refined using a Diffusion model. These sequential manipulations, referred to as \textit{multi-step deepfakes}, introduce subtle but more complex artifacts with semantic shifts that can possibly and significantly evade the existing conventional forensic analysis.


Thus, we believe that multi-step deepfake creation pipelines introduce critical yet underexplored challenges for the existing detectors. First, multi-step forgeries are not simply additive, but involve complex, nonlinear interactions across spectral, spatial, and semantic domains. For instance, one generator may suppress or overwrite artifacts introduced by a previous step, making it considerably harder to trace manipulation history. In practice, deepfake content may undergo multiple transformations such as identity swapping, aging, relighting, or style enhancement, with each stage progressively obfuscating traces of prior edits~\cite{xu2021facecontroller,li2016deep}. Second, as synthetic data becomes increasingly widespread~\cite{nvidia_synthetic_2025,kapania2025examining}, there is a growing risk of recursive use, where generative content from one pipeline is reused in another, often without provenance tracking. This uncontrolled reuse can result in deepfakes being constructed on top of other deepfakes, further complicating detection and forensic attribution. In some cases, this multi-step composition is intentional. For example, an attacker might first apply Face-Swapping and then use GAN-based enhancement or relighting to obscure visible artifacts, making the forgery more realistic and difficult for existing detectors to catch. As a result, real-world deepfake content is likely to contain such composite yet multi-layered manipulations, posing new challenges and threats for existing detectors. 

Unfortunately, such multi-step manipulation trends are severely underestimated in widely used benchmarks or training protocols. Existing detection methods have made considerable progress, particularly for single-step manipulations. Benchmark datasets such as FaceForensics++~\cite{rossler2019faceforensics++} and CelebDF~\cite{CelebDF} have enabled the development of powerful detectors, but they are limited in that they primarily consist of fakes from a single manipulation method.

As a result, current models are predominantly trained to recognize surface-level artifacts introduced by single-step forgeries, such as those in Figure~\ref{fig:singlevsmulti}(a), lacking supervision signals that reflect how these artifacts evolve or interact in multi-step compositions. This poses a serious challenge in forensic settings, where manipulated content often arises from unknown and unstructured generation pipelines. 
In practice, detectors operate under uncertainty, as they lack information about the number and the sequence of specific manipulations applied, as well as the generative origins of the manipulations. Yet, most detectors are implicitly trained under idealized conditions aligned with in-distribution inference, assuming well-labeled and artifact-consistent data. A detector trained to recognize a specific fingerprint, such as GAN, may perform well when that fingerprint appears in the final output, but can fail when it is suppressed or overwritten by subsequent manipulations~\cite{wesselkamp2022misleading,neves2019ganprintr}. This surface-level reliance severely limits generalization in open-world scenarios, motivating the need for forensic methods that reason beyond visible output and account for compositional manipulation history. Recent work has explored sequential deepfake detection by modeling step-wise manipulations, such as latent edits~\cite{shao2021detecting} or feature trajectories~\cite{xia2023mmnet}, but remains limited to intra-domain transformations and lacks generalization to cross-paradigm pipelines.

We address these challenges by designing a systematic evaluation framework that explicitly considers multi-step manipulation pipelines. Our approach evaluates both homogeneous and heterogeneous manipulations in two-step compositions and focuses on heterogeneous configurations in three-step cases, better reflecting realistic and unstructured synthesis scenarios.
We propose, \textbf{FakeChain}, a benchmark dataset constructed for controlled and reproducible analysis of multi-step deepfakes. As illustrated in Figure~\ref{fig:singlevsmulti}(b), each image in FakeChain is generated by chaining two or three generative models selected from five representative techniques: two GANs (StyleGAN3~\cite{StyleGAN3}, and StyleSwin~\cite{StyleSwin}), two Diffusion models (Stable Diffusion 3~\cite{StableDiffusion3}, and Stable Diffusion XL~\cite{Podell2023}), and one Face-Swapping method (FaceFusion~\cite{facefusion}). This design enables systematic analysis of how generation depth and manipulation sequence influence spectral characteristics, feature representations, and detection performance.

Through extensive experiments and empirical analysis, we find that existing detectors fail to generalize over FakeChain, and we uncover that the final manipulation step is the most important for detection.  

Our contributions are summarized as follows:

\begin{enumerate}[leftmargin=10pt]
    \item \textbf{FakeChain: A Benchmark for Multi-Step Deepfakes.} We introduce FakeChain, the first benchmark dataset of manipulated face images synthesized through 2- and 3-Step fake generation chains across five representative techniques. Each sample is provided with prompt metadata info and manipulation history, enabling reproducible and interpretable forensic analysis.

    \item \textbf{Comprehensive Analysis of Multi-Step Manipulations.} We conduct feature-level (t-SNE), spectral (FFT), and mutual information analyses with detector benchmarking to assess how manipulation depth, final-step generator, and composition affect detectability. Our findings reveal distinct patterns of frequency overwriting and feature collapse based on the final generator.

    \item \textbf{Revealing Final-Step Bias in Multi-Step Detection.} We show that current detectors are heavily biased toward final-step artifacts, failing to account for earlier manipulations. This bias leads to severe performance drops (up to 58.83\% F1-score) when the final generator differs from training, highlighting the urgent need for multi-step benchmarks such as FakeChain to expose such compositional vulnerabilities.

\end{enumerate}


\section{Related Work}

\subsection{Deepfake Generation}

\textsc{\textbf{Face-swapping Methods.}} Face-swapping aims to synthesize images in which a source identity is transferred onto a target face~\cite{korshunova2017fast}, preserving the pose~\cite{rosberg2023facedancer}, lighting~\cite{he2025high}, and expression of the target~\cite{Chen2020SimSwap}, while replacing its identity with that of the source. This task inherently involves two inputs: a source image, representing the identity to be transferred, and a target image, whose structure and context are preserved.
Earlier work, such as the autoencoder-based approach~\cite{faceswap2018}, relied on identity disentanglement through shared and separate encoder-decoder pairs. These models laid the foundation for source-target identity translation. However, it lacked spatial fidelity and robustness. Subsequent GAN-based models advanced the field. \textit{FaceShifter}~\cite{Li2020FaceShifter} introduced a two-stage pipeline: first, a coarse face swap via global identity embedding. Then, a refinement stage guided by spatial attention to resolve occlusions and misalignments. \textit{SimSwap}~\cite{Chen2020SimSwap} further generalized the process by introducing an identity injection module that flexibly supports arbitrary source-target pairings without additional training.
More recently, Diffusion-based face-swapping approaches have emerged. \textit{DiffSwap}~\cite{Zhou2023DiffSwap} formulates face-swapping as a conditional denoising problem, using 3D face alignment to guide identity injection during the Diffusion process. This method achieves high perceptual quality while maintaining target fidelity and source identity preservation.

\noindent\textsc{\textbf{GAN-based Face Synthesis.}} Generative Adversarial Networks (GANs) have significantly advanced facial image synthesis, transitioning from low-resolution outputs to high-fidelity, photorealistic images. Early influential works include \textit{ProGAN}~\cite{karras2018progressive}, which pioneered progressive growing of networks, enabling stable training at unprecedented resolutions by incrementally adding network layers. Following ProGAN's success, style-based architectures such as \textit{StyleGAN}~\cite{Karras2019StyleGAN} introduced a novel latent-space manipulation mechanism, facilitating controlled edits of facial attributes and styles. Subsequent refinements in \textit{StyleGAN2}~\cite{Karras2020StyleGAN2} improved upon image quality by addressing inherent GAN artifacts such as texture sticking and phase issues. More recently, \textit{StyleGAN3}~\cite{Karras2021AliasFreeGAN} proposed alias-free generation methods to eliminate subtle spatial aliasing artifacts and enhance rotation equivariance.
Beyond the StyleGAN family, recent models such as \textit{GigaGAN}~\cite{kang2023gigagan} further pushed boundaries in scalability and resolution, employing multi-scale training and feature disentanglement to produce diverse, high-resolution facial images efficiently. These advances collectively underscore GANs' critical role in deepfake generation, influencing both the synthesis quality and the complexity of forensic detection tasks.


\noindent\textsc{\textbf{Diffusion-based Facial Synthesis.}}
Diffusion models have recently emerged as state-of-the-art in high-fidelity image generation, particularly for face synthesis. Early methods such as Denoising Diffusion Probabilistic Models (DDPM)~\cite{ho2020ddpm} demonstrated that realistic samples could be produced through iterative noise refinement. OpenAI's \textit{GLIDE}~\cite{nichol2021glide} extended this framework by introducing text-conditional generation using CLIP embeddings, and proposed \textit{classifier-free guidance} (CFG) to improve sample controllability without relying on external classifiers.
This guidance mechanism was widely adopted in subsequent models, most notably the \textit{Latent Diffusion Model (LDM)}~\cite{rombach2022ldm}, which performed denoising in a compressed latent space to significantly reduce computational costs. LDM laid the foundation for \textit{Stable Diffusion}, an open-source Diffusion model that made high-quality synthesis accessible at scale. Its successor, \textit{Stable Diffusion XL (SDXL)}~\cite{Podell2024SDXL}, further improved output fidelity and expressiveness, supporting diverse face manipulation tasks.
The iterative nature and flexible conditioning of Diffusion models make them especially well-suited for multi-step facial synthesis. They support identity-preserving generation, semantic attribute control, and high-resolution refinement, all of which are key requirements in complex deepfake workflows.

\subsection{Deepfake Detection}

Deepfake detection methods have progressed from early CNN-based models to more advanced architectures incorporating transformers and frequency cues. Xception~\cite{chollet2017xception} serves as a strong CNN baseline for spatial feature extraction. More recent approaches such as CCViT~\cite{coccomini2022combining} combine EfficientNet with Vision Transformer modules to enhance robustness. Transformer-based models, for example ViViT~\cite{arnab2021vivit} and Swin Transformer~\cite{liu2021swin} are also leveraged to capture long-range and hierarchical features in complex manipulations. MAT~\cite{DBLP:journals/corr/abs-2103-02406} introduces multi-attentional fusion, improving resilience against local distortions.

Various approaches~\cite{4,6,8,25,28,29,pei2024deepfake} have also been proposed to enhance the robustness of deepfake detectors. One line of work focuses on transforming the input data domain to exploit spectral cues; for example, F3Net~\cite{qian2020thinking} incorporates frequency-aware modules into its architecture, while Durall~\textit{et al.}~\cite{durall2020watch} and Frank\textit{et al.}~\cite{frank2020leveraging} show that generative models leave distinctive traces in the frequency domain. Another line of research introduces unique training paradigms to improve generalization: DiffusionFake\cite{chen2024diffusionfake} targets generator-specific artifacts, ADD~\cite{le2021add} formulates detection as an anomaly detection task without relying on known forgeries, and QAD~\cite{Le_2023_ICCV} enhances robustness under varying compression and resolution. While these models show strong performance in their respective settings, they are still primarily benchmarked on single-step manipulation. As manipulation pipelines become more compositional, there is a growing need for detectors that consider manipulation history and sequence rather than only surface-level artifacts.

\subsection{Generalization and Compositional Manipulation}
Despite the significant progress in deepfake detection, generalization remains a considerable challenge. Most existing detectors perform strongly on seen manipulations but often degrade significantly under novel or compositional scenarios.
Recent work has begun exploring sequential manipulation detection. For instance, SeqFakeFormer~\cite{shao2021detecting} frames multi-step forgery analysis as an image-to-sequence prediction task, while MMNet~\cite{xia2023mmnet} introduces a multi-branch architecture to recover manipulation regions across sequentially altered faces. These approaches mark an important first step toward compositional deepfake analysis. However, both methods remain limited in two critical aspects: they operate exclusively within GAN-based latent spaces, and they assume access to consistent manipulation patterns. As such, they fall short in addressing the growing number of real-world scenarios where multi-step deepfakes are constructed by chaining diverse generation paradigms, such as Face-Swapping, GAN synthesis, and Diffusion-based refinement.

In contrast, our work is the first to systematically analyze multi-step manipulations involving heterogeneous generators. Leveraging a newly constructed benchmark, we study how manipulation depth and composition affect detectability and spectral traces, uncovering overlooked biases in existing detection models.
\section{Methodology}

\subsection{Deepfake Generation Methods}


\textsc{\textbf{Face-swapping Method.}}
We adopt \textbf{\textit{FaceFusion}}~\cite{facefusion}, a modular image-based face-swapping framework that supports reference-guided alignment, semantic mask blending, and fine-grained identity control. Unlike early autoencoder-based approaches such as FaceSwap~\cite{faceswap2018}, FaceFusion separates the face-swapping pipeline into four discrete stages: face detection, landmark alignment, identity embedding, and semantic-aware blending.

This modular design allows high-quality facial replacement by preserving spatial structure and ensuring natural boundary transitions. Particularly for static image synthesis, FaceFusion offers precise control over face alignment and identity matching, enabling reliable and consistent manipulation suitable for multi-step deepfake generation pipelines.

\noindent\textsc{\textbf{GAN-based Generation.}} For GAN-based synthesis, we employ \textbf{\textit{StyleGAN3}}~\cite{StyleGAN3}, which introduces alias-free convolutions to generate temporally coherent and spatially stable images, and \textbf{\textit{StyleSwin}}~\cite{StyleSwin}, which leverages a transformer-based generator architecture for improved semantic consistency and high-resolution synthesis. These models have demonstrated strong capabilities in synthesizing photorealistic human faces from latent representations, which has led to their frequent use as generative backbones or prior models in face synthesis pipelines. Their architectural diversity also helps evaluate the robustness of detectors against stylistic and structural variance in GAN-based forgeries.

\noindent\textsc{\textbf{Diffusion-based Generation.}} For Diffusion-based synthesis, we include \textbf{\textit{Stable Diffusion 3}}~\cite{StableDiffusion3} and \textbf{\textit{Stable Diffusion XL}}~\cite{Podell2023}, both of which implement classifier-free guidance and latent-space denoising for efficient and controllable generation. Stable Diffusion 3 is known for balancing generation speed and quality, while Stable Diffusion XL enhances resolution and semantic control in fine-grained face synthesis. Their iterative refinement mechanisms make them especially well-suited for multi-step manipulation pipelines, where conditional generation is repeatedly applied.

\subsection{Multi-Step Deepfake Dataset Construction}

With the goal of systematically investigating how sequential manipulations affect image characteristics and detectability, we construct FakeChain, a benchmark dataset of one-step, two-step, and three-step manipulated face images. All manipulations are drawn from five distinct generation techniques: one face-swapping method (FaceFusion~\cite{facefusion}), two GAN-based models (StyleGAN3~\cite{StyleGAN3} and StyleSwin~\cite{StyleSwin}), and two Diffusion-based models (Stable Diffusion 3~\cite{StableDiffusion3} and Stable Diffusion XL~\cite{Podell2023}). These methods represent three distinct generative paradigms—face-swapping, GAN-based synthesis, and Diffusion-based refinement—covering a broad range of deepfake creation strategies.
All real images in FakeChain originate from the FFHQ-1024 dataset~\cite{ffhq}. This high-quality facial image collection serves as the foundation for face-swapping operations and as a reference for evaluation tasks.

The generation pipeline and modality-specific input configurations for constructing these multi-step manipulations are summarized in ~\autoref{tab:step_generation}. This framework outlines how different input modalities, such as latent vectors, text prompts, and real face images, are used across 1-Step, 2-Step, and 3-Step manipulations, respectively, depending on the generation method. 



\subsection{\textbf{Notation for Manipulation Chains}} 
First, we denote the term as $x_{\text{fake}}=\mathcal{G}_i(x_j)$, where $x_{\text{fake}}$ is the deepfake output generated from the generation method $\mathcal{G}$ with $i$th step with input image $x_j$ (See Table 1). Also, we define two complementary notations: one for manipulation chains and the other for evaluation grouping, in order to maintain clarity and consistency across generation and evaluation.

\textbf{\textbf{- Generation Chain Notation.}} 
We denote the full sequence of generators using shorthand tokens: \textbf{FF} (FaceFusion), \textbf{SG3} (StyleGAN3), \textbf{SS} (StyleSwin), \textbf{SD3} (Stable Diffusion 3), and \textbf{SDXL} (Stable Diffusion XL). Sequential manipulations are joined by underscores; for example, \texttt{FF\_SG3} indicates FaceFusion followed by StyleGAN3.

\textbf{\textbf{- Evaluation Grouping Notation.}}
To simplify performance comparisons, we group samples by final generator type and manipulation depth: \textbf{1-Step FS}, \textbf{1-Step GAN}, \textbf{1-Step Diff} for single-step fakes, and \textbf{2-Step} or \textbf{3-Step FS/GAN/Diff} for multi-step fakes categorized by their final generator. This reflects the dominant influence of the last manipulation step during evaluation.


\begin{table}[t]
\small
\setlength{\tabcolsep}{4pt}
\renewcommand{\arraystretch}{1.1}
\centering

\begin{tabular}{p{1.6cm} p{6.5cm}}
\toprule
\textbf{Step} & \textbf{Generation Process} \\
\midrule
1-Step 
& $x_{\text{fake}} = \mathcal{G}_1(x_0)$ \newline
- If $\mathcal{G}_1 = \text{FF}$: the input $x_0 = \{s, t\}$ consists of a source image $s$ and a target image $t$, which are randomly sampled from real images (e.g., FFHQ-1024).
\newline
- If $\mathcal{G}_1 \in \{\text{SG3}, \text{SS}\}$: $x_0 = z$, random latent vector. \newline
- If $\mathcal{G}_1 \in \{\text{SD3}, \text{SDXL}\}$: $x_0 = p$, text prompt. \newline
- All 5 individual generators are used.
\\

\midrule
2-Step 
& $x_{\text{fake}} = \mathcal{G}_2(\mathcal{G}_1(x_0))$ \newline
- If $\mathcal{G}_2 = \text{FF}$: target is $\mathcal{G}_1(x_0)$, source $s'$ is sampled from real images, FFHQ-1024. \newline
- If $\mathcal{G}_2 \in \{\text{SG3}, \text{SS}\}$: latent is extracted from $\mathcal{G}_1(x_0)$. \newline
- If $\mathcal{G}_2 \in \{\text{SD3}, \text{SDXL}\}$: prompt $p$ is used with $\mathcal{G}_1(x_0)$ as conditioning image. \newline
- All 20 generator pairs are used (including both homogeneous and heterogeneous combinations). \\
\midrule
3-Step 
& $x_{\text{fake}} = \mathcal{G}_3(\mathcal{G}_2(\mathcal{G}_1(x_0)))$ \newline
- If $\mathcal{G}_3 = \text{FF}$: target is $\mathcal{G}_2(\mathcal{G}_1(x_0))$, and source $s''$ is sampled from FFHQ-1024. \newline
- If $\mathcal{G}_3 \in \{\text{SG3}, \text{SS}\}$: latent is extracted from $\mathcal{G}_2(\mathcal{G}_1(x_0))$. \newline
- If $\mathcal{G}_3 \in \{\text{SD3}, \text{SDXL}\}$: prompt $p'$ is used with \newline $\mathcal{G}_2(\mathcal{G}_1(x_0))$ as conditioning image.  \newline
- All 24 generator chains are heterogeneous combinations (no repetition of the same generator type). \\
\bottomrule

\end{tabular}
\caption{Generation pipeline and modality-specific input configurations for constructing 1-Step, 2-Step, and 3-Step deepfakes in the FakeChain.}
\label{tab:step_generation}
\end{table}


\subsection{\textbf{Step-wise Manipulation Process}}
\subsubsection{\textbf{Step-One Manipulations (1-Step).}} The one-Step subset consists of five manipulation types, each reflecting the output of a single generator. For the \textit{FaceFusion} case, fake images are created by randomly pairing source and target faces within the FFHQ-1024 dataset. This design ensures that the resulting forgeries reflect only the identity transfer performed by the face-swapping operation.

In contrast, the GAN-based (StyleGAN3, StyleSwin) and Diffusion-based (Stable Diffusion 3, Stable Diffusion XL) images are synthesized independently, without using FFHQ-1024 images as input. The GAN models generate faces from random latent codes, while the Diffusion models are conditioned on textual prompts to guide semantic synthesis. These samples represent deepfakes originating from generative priors rather than real-image conditioning.

For the Diffusion-based samples, prompts were randomly generated from the full Cartesian product of age, ethnicity, and gender attributes, as shown below:

\begin{tcolorbox}[title=Prompt Attribute Sets, colback=gray!10, colframe=gray!40, boxrule=0.5pt, arc=2pt, left=4pt, right=4pt, top=4pt, bottom=4pt, title style={color=black}]
\begin{itemize}[leftmargin=1.2em, label={}]
    \item \textbf{Ages}: \texttt{["5", "10", "15", "20", "25", "30", "40", "50", "60", "70", "80"]}
    \item \textbf{Ethnicities}: \texttt{["Caucasian", "Middle Eastern", "Asian", "African", "Hispanic", "South Asian", "Native American", "Pacific Islander"]}
    \item \textbf{Genders}: \texttt{["man", "woman"]}
\end{itemize}
\end{tcolorbox}




\subsubsection{\textbf{Step-Two Manipulations (2-Step).}} 
We generate all pairwise combinations among the five manipulation methods, resulting in 20 two-Step configurations, including both \textit{homogeneous} sequences such as GAN$\rightarrow$GAN and \textit{heterogeneous} sequences such as GAN$\rightarrow$Diffusion. When \textit{FaceFusion} is applied in the second step, the output of the first manipulation serves as the target, while a new source face is sampled from FFHQ-1024, maintaining identity transfer grounded in real facial features.


\subsubsection{\textbf{Step-Three Manipulations (3-Step).}} 
The three-Step subset includes 24 heterogeneous manipulation chains. Homogeneous combinations are excluded to prioritize realistic and diverse compositional scenarios. As in the two-Step setting, when \textit{FaceFusion} is used at any stage, the source is drawn from FFHQ-1024 and the target is always the output of the preceding step.


\subsection{\textbf{Metadata and Reproducibility}} 
For all images involving \textit{FaceFusion}, we record the corresponding source--target identity pair. For two- and three-Step forgeries, we additionally log the full manipulation sequence, including the original input image (if applicable), the order of applied manipulations, and the text prompts used in Diffusion stages. This metadata is saved in structured text files alongside the generated images, facilitating traceable analysis and reproducibility.

\section{Experiments}

\subsection{Evaluation Protocols and Metrics}

We adopt a two-pronged evaluation strategy to assess the detectability and representation dynamics of multi-step deepfakes, combining detection through classification models with frequency-domain analysis. Specifically, we benchmark three representative detection models and conduct spectral inspection using FFT-based analysis.

The FakeChain dataset comprises 49 unique combinations of training and testing setups, reflecting all possible pairings among 1-Step, 2-Step, and 3-Step manipulation types involving different final generators. For each manipulation combination, we generate 25{,}000 images and split them into training, validation, and test subsets using a fixed 70\%, 20\%, and 10\% ratio respectively. This consistent partitioning ensures fair comparisons across models and manipulation types.

We report standard binary classification metrics including accuracy, F1-score, and AUC to quantify detector performance. Additionally, t-SNE is used to visualize penultimate-layer features, providing insight into how multi-step forgeries are represented in the learned embedding space.





\subsection{Deepfake Detection Models}


We benchmark three representative deepfake detection models that span complementary architectural paradigms of convolutional, frequency-aware, and attention-based learning. Xception serves as a widely recognized CNN baseline and has been extensively adopted in prior benchmarks such as FaceForensics++, making it an indispensable reference for spatial feature extraction. F3Net represents frequency-aware detectors, explicitly leveraging spectral artifacts that are highly relevant to the study of how multi-step manipulations progressively distort or erase frequency signals. MAT exemplifies transformer-based approaches, introducing multi-attentional fusion that captures both local and global artifacts and provides resilience against diverse distortions. These models were also chosen for their publicly available implementations and pretrained weights, ensuring reproducibility and fair comparison within the proposed FakeChain benchmark.


All models are trained on the FakeChain dataset using six distinct training configurations, which include 1-Step and 2-Step variants for each of the three generator types (FS, GAN, and Diff). Evaluation is performed on all nine categories, consisting of 1-Step, 2-Step, and 3-Step data for each generator type, resulting in a total of 54 unique train-test combinations. Fine-tuning is performed from publicly available pretrained weights for up to 10 epochs, with early stopping based on validation accuracy (patience = 3). Training is conducted independently for each combination to isolate the effect of manipulation depth and type. Other training hyperparameters, such as optimizer settings and learning rate schedules, follow the default configurations provided in each model’s original implementation.


\subsection{Feature Space Visualization}

We employ t-SNE~\cite{van2008visualizing} to visualize penultimate-layer embeddings from each detection model, enabling qualitative analysis of how multi-step deepfakes are represented in feature space. This projection helps reveal whether different manipulation depths (e.g., 1-, 2-, 3-Step) and generator types result in distinguishable clusters or overlapping distributions. Patterns observed in the embeddings can indicate whether detectors focus primarily on the final manipulation or retain sensitivity to the manipulation history.

By examining the spatial organization of these embeddings, we gain insight into each model's internal decision-making behavior, including its susceptibility to surface-level artifacts and capacity for capturing compositional structures.

\subsection{Frequency Spectrum Analysis}
We perform an FFT-based analysis to examine the spectral characteristics of multi-step deepfakes and identify manipulation-specific artifacts in the high-frequency domain~\cite{durall2020watch}. Each image is converted to grayscale and high-pass filtered by subtracting a median-blurred version, suppressing low-frequency structures while retaining generative artifacts. 

The magnitude spectrum is computed from the high-pass image using the 2D discrete Fourier transform (FFT), followed by frequency shifting and logarithmic scaling for improved interpretability. This analysis enables direct comparison of spectral signatures across manipulation depths and techniques, and helps assess whether the final-stage generator dominates the frequency characteristics of multi-step forgeries.

\begin{table*}[t]
\centering
\resizebox{\textwidth}{!}{%
\begin{tabular}{l|l|l|llllll|llllll|llllll}
\hline
 & \multicolumn{1}{c|}{} & \multicolumn{1}{c|}{} & \multicolumn{6}{c|}{\textbf{1- Step}} & \multicolumn{6}{c|}{\textbf{2-Step}} & \multicolumn{6}{c}{\textbf{3-Step}} \\
 & \multicolumn{1}{c|}{} & \multicolumn{1}{c|}{} & \multicolumn{2}{c}{\textit{FS}} & \multicolumn{2}{c}{\textit{GAN}} & \multicolumn{2}{c|}{\textit{Diff}} & \multicolumn{2}{c}{\textit{FS}} & \multicolumn{2}{c}{\textit{GAN}} & \multicolumn{2}{c|}{\textit{Diff}} & \multicolumn{2}{c}{\textit{FS}} & \multicolumn{2}{c}{\textit{GAN}} & \multicolumn{2}{c}{\textit{Diff}} \\ \cline{4-21} 
 & \multicolumn{1}{c|}{\multirow{-3}{*}{\textbf{\textsc{Detectors}}}} & \multicolumn{1}{c|}{\multirow{-3}{*}{\textbf{\textsc{Train on}}}} & \multicolumn{1}{c}{AUC} & \multicolumn{1}{c}{F1} & \multicolumn{1}{c}{AUC} & \multicolumn{1}{c}{F1} & \multicolumn{1}{c}{AUC} & \multicolumn{1}{c|}{F1} & \multicolumn{1}{c}{AUC} & \multicolumn{1}{c}{F1} & \multicolumn{1}{c}{AUC} & \multicolumn{1}{c}{F1} & \multicolumn{1}{c}{AUC} & \multicolumn{1}{c|}{F1} & \multicolumn{1}{c}{AUC} & \multicolumn{1}{c}{F1} & \multicolumn{1}{c}{AUC} & \multicolumn{1}{c}{F1} & \multicolumn{1}{c}{AUC} & \multicolumn{1}{c}{F1} \\ \cline{2-21} 
 &  & \textit{FS} & \cellcolor[HTML]{D9D9D9}\textbf{99.92} & \cellcolor[HTML]{D9D9D9}\textbf{99.35} & 49.65 & 0.44 & 59.49 & 21.98 & \cellcolor[HTML]{D9D9D9}99.37 & \cellcolor[HTML]{D9D9D9}96.47 & 50.20 & 0.35 & 60.51 & 9.44 & \cellcolor[HTML]{D9D9D9}\textbf{99.99} & \cellcolor[HTML]{D9D9D9}\textbf{99.71} & 49.25 & 0.30 & 61.56 & 5.33 \\
 &  & \textit{GAN} & 59.77 & 8.11 & \cellcolor[HTML]{D9D9D9}99.95 & \cellcolor[HTML]{D9D9D9}98.99 & 55.47 & 6.23 & 70.51 & 24.35 & \cellcolor[HTML]{D9D9D9}99.95 & \cellcolor[HTML]{D9D9D9}99.02 & 61.29 & 4.91 & 73.16 & 27.49 & \cellcolor[HTML]{D9D9D9}99.97 & \cellcolor[HTML]{D9D9D9}99.20 & 66.82 & 6.63 \\
 & \multirow{-3}{*}{Xception} & \textit{Diff} & 79.30 & 0.55 & 49.39 & 0.04 & \cellcolor[HTML]{D9D9D9}\textbf{100.00} & \cellcolor[HTML]{D9D9D9}\textbf{100.00} & 90.28 & 41.46 & 44.65 & 0.03 & \cellcolor[HTML]{D9D9D9}\textbf{99.72} & \cellcolor[HTML]{D9D9D9}\textbf{88.56} & 86.89 & 6.66 & 46.16 & 0.13 & \cellcolor[HTML]{D9D9D9}\textbf{99.57} & \cellcolor[HTML]{D9D9D9}\textbf{79.65} \\ \cline{2-21} 
 &  & \textit{FS} & \cellcolor[HTML]{D9D9D9}98.86 & \cellcolor[HTML]{D9D9D9}98.86 & 50.00 & 1.50 & 55.14 & 19.66 & \cellcolor[HTML]{D9D9D9}97.38 & \cellcolor[HTML]{D9D9D9}97.25 & 49.97 & 1.35 & 54.83 & 17.44 & \cellcolor[HTML]{D9D9D9}99.40 & \cellcolor[HTML]{D9D9D9}99.40 & 50.30 & 2.54 & 56.40 & 22.02 \\
 &  & \textit{GAN} & 69.92 & 59.83 & \cellcolor[HTML]{D9D9D9}96.42 & \cellcolor[HTML]{D9D9D9}96.47 & 58.71 & 34.77 & 73.80 & 65.34 & \cellcolor[HTML]{D9D9D9}94.82 & \cellcolor[HTML]{D9D9D9}94.39 & 56.24 & 27.58 & 80.85 & 77.12 & \cellcolor[HTML]{D9D9D9}95.13 & \cellcolor[HTML]{D9D9D9}95.04 & 57.33 & 30.82 \\
 & \multirow{-3}{*}{F3Net} & \textit{Diff} & 50.00 & 0.08 & 49.98 & 0.00 & \cellcolor[HTML]{D9D9D9}99.98 & \cellcolor[HTML]{D9D9D9}99.98 & 69.51 & 43.83 & 49.98 & 0.01 & \cellcolor[HTML]{D9D9D9}82.93 & \cellcolor[HTML]{D9D9D9}77.06 & 50.33 & 1.37 & 50.00 & 0.09 & \cellcolor[HTML]{D9D9D9}75.65 & \cellcolor[HTML]{D9D9D9}66.28 \\ \cline{2-21} 
 &  & \textit{FS} & \cellcolor[HTML]{D9D9D9}99.17 & \cellcolor[HTML]{D9D9D9}98.81 & 45.71 & 0.52 & 71.73 & 13.71 & \cellcolor[HTML]{D9D9D9}\textbf{99.56} & \cellcolor[HTML]{D9D9D9}\textbf{98.49} & 42.26 & 0.45 & 53.94 & 9.78 & \cellcolor[HTML]{D9D9D9}99.74 & \cellcolor[HTML]{D9D9D9}99.55 & 34.05 & 0.37 & 51.09 & 13.90 \\
 &  & \textit{GAN} & 51.98 & 0.08 & \cellcolor[HTML]{D9D9D9}\textbf{100.00} & \cellcolor[HTML]{D9D9D9}\textbf{100.00} & 91.34 & 0.12 & 72.03 & 1.71 & \cellcolor[HTML]{D9D9D9}\textbf{100.00} & \cellcolor[HTML]{D9D9D9}\textbf{100.00} & 51.13 & 0.00 & 47.97 & 0.66 & \cellcolor[HTML]{D9D9D9}\textbf{100.00} & \cellcolor[HTML]{D9D9D9}\textbf{100.00} & 42.29 & 0.01 \\
\multirow{-12}{*}{\rotatebox[origin=c]{90}{\textbf{1-Step}}} & \multirow{-3}{*}{MAT} & \textit{Diff} & 65.43 & 0.32 & 50.68 & 0.16 & \cellcolor[HTML]{D9D9D9}\textbf{100.00} & \cellcolor[HTML]{D9D9D9}\textbf{100.00} & 82.91 & 49.15 & 53.29 & 0.24 & \cellcolor[HTML]{D9D9D9}99.54 & \cellcolor[HTML]{D9D9D9}82.71 & 72.49 & 1.85 & 47.71 & 0.59 & \cellcolor[HTML]{D9D9D9}99.29 & \cellcolor[HTML]{D9D9D9}70.78 \\ \hline
\end{tabular}
}
\caption{
Detection performance of models trained on \textbf{1-Step} manipulated data and evaluated across 1-, 2-, and 3-Step test sets, grouped by final manipulation type. Each cell reports both AUC and F1-score. \textbf{Bold} values denote the best-performing model for each test case, while \cellcolor[HTML]{D9D5D5}gray cells indicate cases where the final manipulation method matches the training configuration.
}
\label{tab:stepwise_detection1}
\end{table*}

\begin{table*}[t]
\centering
\resizebox{\textwidth}{!}{%
\begin{tabular}{l|l|l|llllll|llllll|llllll}
\hline
 & \multicolumn{1}{c|}{} & \multicolumn{1}{c|}{} & \multicolumn{6}{c|}{\textbf{1-Step}} & \multicolumn{6}{c|}{\textbf{2-Step}} & \multicolumn{6}{c}{\textbf{3-Step}} \\
 & \multicolumn{1}{c|}{} & \multicolumn{1}{c|}{} & \multicolumn{2}{c}{\textit{FS}} & \multicolumn{2}{c}{\textit{GAN}} & \multicolumn{2}{c|}{\textit{Diff}} & \multicolumn{2}{c}{\textit{FS}} & \multicolumn{2}{c}{\textit{GAN}} & \multicolumn{2}{c|}{\textit{Diff}} & \multicolumn{2}{c}{\textit{FS}} & \multicolumn{2}{c}{\textit{GAN}} & \multicolumn{2}{c}{\textit{Diff}} \\ \cline{4-21} 
 & \multicolumn{1}{c|}{\multirow{-3}{*}{\textbf{\textsc{Detectors}}}} & \multicolumn{1}{c|}{\multirow{-3}{*}{\textbf{Train on}}} & \multicolumn{1}{c}{AUC} & \multicolumn{1}{c}{F1} & \multicolumn{1}{c}{AUC} & \multicolumn{1}{c}{F1} & \multicolumn{1}{c}{AUC} & \multicolumn{1}{c|}{F1} & \multicolumn{1}{c}{AUC} & \multicolumn{1}{c}{F1} & \multicolumn{1}{c}{AUC} & \multicolumn{1}{c}{F1} & \multicolumn{1}{c}{AUC} & \multicolumn{1}{c|}{F1} & \multicolumn{1}{c}{AUC} & \multicolumn{1}{c}{F1} & \multicolumn{1}{c}{AUC} & \multicolumn{1}{c}{F1} & \multicolumn{1}{c}{AUC} & \multicolumn{1}{c}{F1} \\ \cline{2-21} 
 &  & \textit{FS} & \cellcolor[HTML]{D9D9D9}\textbf{99.46} & \cellcolor[HTML]{D9D9D9}\textbf{99.15} & 87.36 & 4.61 & 98.24 & 85.84 & \cellcolor[HTML]{D9D9D9}99.99 & \cellcolor[HTML]{D9D9D9}\textbf{99.91} & 92.27 & 8.69 & 88.20 & 32.80 & \cellcolor[HTML]{D9D9D9}\textbf{99.99} & \cellcolor[HTML]{D9D9D9}\textbf{99.91} & 82.54 & 6.24 & 89.42 & 19.60 \\
 &  & \textit{GAN} & 40.58 & 0.63 & \cellcolor[HTML]{D9D9D9}96.17 & \cellcolor[HTML]{D9D9D9}68.02 & 55.32 & 1.30 & 52.93 & 3.32 & \cellcolor[HTML]{D9D9D9}97.64 & \cellcolor[HTML]{D9D9D9}87.45 & 61.83 & 4.28 & 64.50 & 26.19 & \cellcolor[HTML]{D9D9D9}92.31 & \cellcolor[HTML]{D9D9D9}56.25 & 68.59 & 7.38 \\
 & \multirow{-3}{*}{Xception} & \textit{Diff} & 92.62 & 30.14 & 52.67 & 0.44 & \cellcolor[HTML]{D9D9D9}99.99 & \cellcolor[HTML]{D9D9D9}99.74 & 96.52 & 60.21 & 55.25 & 0.77 & \cellcolor[HTML]{D9D9D9}99.99 & \cellcolor[HTML]{D9D9D9}99.73 & 97.02 & 60.02 & 60.16 & 0.94 & \cellcolor[HTML]{D9D9D9}99.99 & \cellcolor[HTML]{D9D9D9}99.74 \\ \cline{2-21} 
 &  & \textit{FS} & \cellcolor[HTML]{D9D9D9}98.80 & \cellcolor[HTML]{D9D9D9}98.79 & 50.02 & 0.24 & 82.57 & 77.60 & \cellcolor[HTML]{D9D9D9}99.71 & \cellcolor[HTML]{D9D9D9}99.71 & 50.03 & 0.28 & 60.43 & 28.39 & \cellcolor[HTML]{D9D9D9}99.78 & \cellcolor[HTML]{D9D9D9}99.78 & 50.07 & 0.42 & 55.27 & 17.63 \\
 &  & \textit{GAN} & 59.90 & 40.87 & \cellcolor[HTML]{D9D9D9}59.64 & \cellcolor[HTML]{D9D9D9}39.64 & 52.26 & 20.57 & 58.63 & 36.84 & \cellcolor[HTML]{D9D9D9}94.75 & \cellcolor[HTML]{D9D9D9}94.87 & 53.56 & 24.06 & 74.46 & 63.43 & \cellcolor[HTML]{D9D9D9}69.15 & \cellcolor[HTML]{D9D9D9}50.11 & 66.85 & 52.22 \\
 & \multirow{-3}{*}{F3Net} & \textit{Diff} & 51.74 & 8.36 & 50.02 & 1.88 & \cellcolor[HTML]{D9D9D9}99.53 & \cellcolor[HTML]{D9D9D9}99.53 & 62.03 & 35.63 & 49.96 & 1.62 & \cellcolor[HTML]{D9D9D9}99.48 & \cellcolor[HTML]{D9D9D9}99.48 & 57.98 & 27.53 & 49.99 & 1.75 & \cellcolor[HTML]{D9D9D9}99.46 & \cellcolor[HTML]{D9D9D9}99.46 \\ \cline{2-21} 
 &  & \textit{FS} & \cellcolor[HTML]{D9D9D9}98.03 & \cellcolor[HTML]{D9D9D9}98.50 & 99.98 & 99.09 & 99.25 & 93.30 & \cellcolor[HTML]{D9D9D9}\textbf{100.00} & \cellcolor[HTML]{D9D9D9}99.80 & 100.0 & 99.73 & 87.59 & 41.02 & \cellcolor[HTML]{D9D9D9}99.92 & \cellcolor[HTML]{D9D9D9}99.63 & 99.99 & 99.66 & 88.42 & 26.84 \\
 &  & \textit{GAN} & 76.99 & 0.32 & \cellcolor[HTML]{D9D9D9}\textbf{100.00} & \cellcolor[HTML]{D9D9D9}\textbf{99.38} & 96.89 & 0.00 & 87.46 & 2.48 & \cellcolor[HTML]{D9D9D9}\textbf{100.00} & \cellcolor[HTML]{D9D9D9}\textbf{100.00} & 74.08 & 0.01 & 76.37 & 1.00 & \cellcolor[HTML]{D9D9D9}\textbf{100.00} & \cellcolor[HTML]{D9D9D9}\textbf{99.61} & 69.56 & 0.05 \\
\multirow{-12}{*}{\rotatebox[origin=c]{90}{\textbf{2-Step}}} & \multirow{-3}{*}{MAT} & \textit{Diff} & 72.59 & 1.90 & 49.34 & 0.12 & \cellcolor[HTML]{D9D9D9}\textbf{100.00} & \cellcolor[HTML]{D9D9D9}\textbf{100.00} & 86.81 & 41.19 & 49.30 & 0.20 & \cellcolor[HTML]{D9D9D9}\textbf{100.00} & \cellcolor[HTML]{D9D9D9}\textbf{100.00} & 79.31 & 4.75 & 56.17 & 0.37 & \cellcolor[HTML]{D9D9D9}\textbf{100.00} & \cellcolor[HTML]{D9D9D9}\textbf{100.00} \\ \hline
\end{tabular}
}
\caption{
Detection performance of models trained on \textbf{2-Step} manipulated data and evaluated across 1-, 2-, and 3-Step test sets, grouped by final manipulation type. Each cell reports both AUC and F1-score. 
\textbf{Bold} values denote the best-performing model for each test case, while \cellcolor[HTML]{D9D5D5}gray cells indicate cases where the final manipulation method matches the training configuration.
}
\label{tab:stepwise_detection2}
\end{table*}

\subsection{Compression Robustness Evaluation with Best Combinations}
We assess detection performance under realistic compression conditions by generating three versions of each test sample: (1) uncompressed \textit{RAW}, (2) \textit{JPEG75}, and (3) \textit{JPEG75x2}, which applies JPEG compression twice at quality level 75. The compression level was chosen with reference to prior research~\cite{dell2025truefake}, which analyzed JPEG quality factors across social media platforms. While Facebook exhibited a broad quality range of 61–92, and X (formerly Twitter) and Telegram used fixed values of 87 and 85, respectively, we selected JPEG 75 as a representative compression level. This value reflects a midpoint within Facebook’s range and applies slightly stronger compression than X and Telegram, enabling evaluation under more challenging conditions. This setup allows evaluation of detector robustness under moderate compression levels that resemble those found in real-world sharing scenarios. Detection was performed using models trained on the best-performing manipulation combinations identified in our earlier analysis.

\section{Results}
\label{sec:results}


\begin{figure*}[t]
  \centering

  \begin{subfigure}[t]{0.32\linewidth}
    \includegraphics[width=\linewidth]{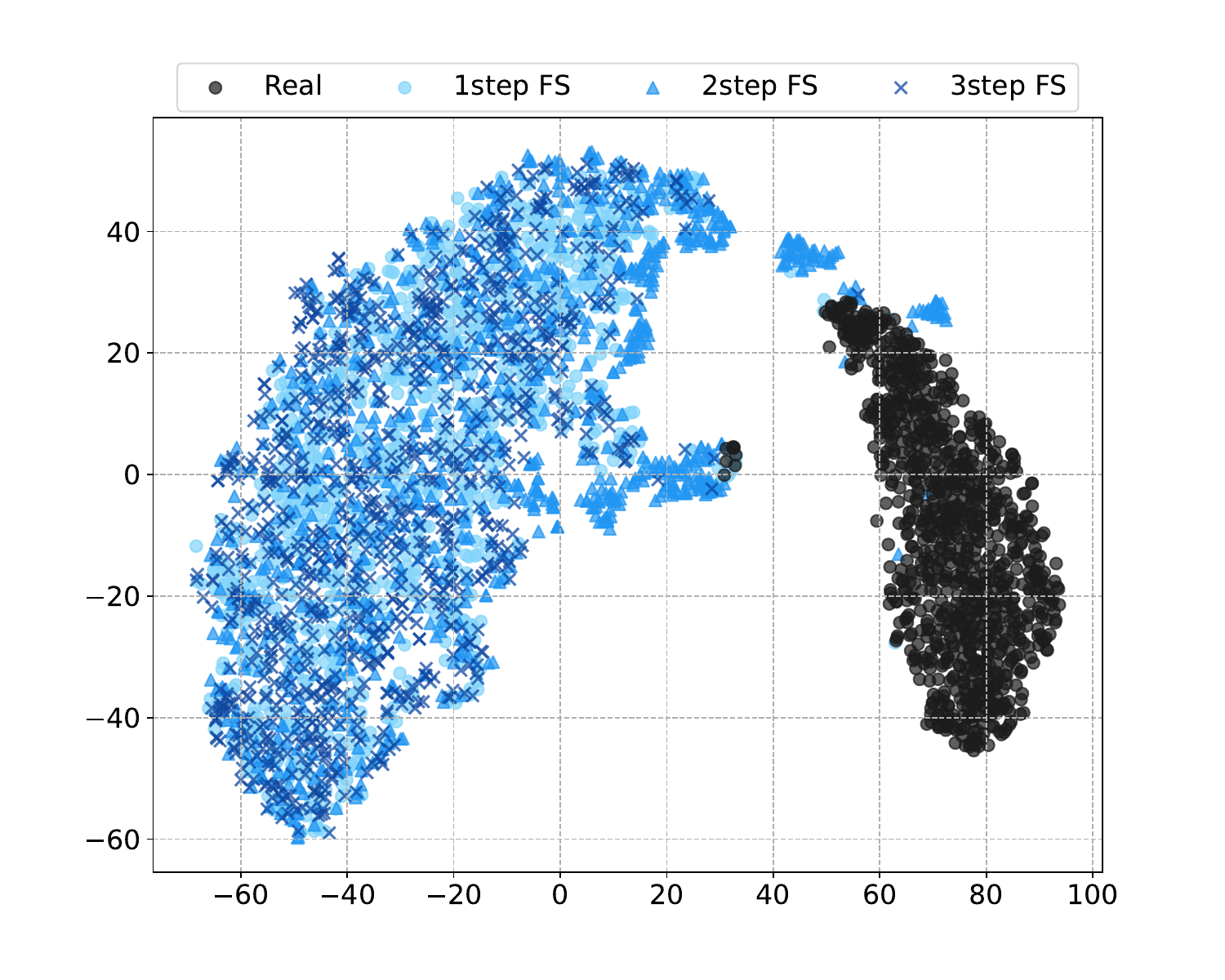}
    \caption{Train: 1-Step FaceSwap}
    \label{fig:tsne_1stepfs}
  \end{subfigure}
  \hfill
  \begin{subfigure}[t]{0.32\linewidth}
    \includegraphics[width=\linewidth]{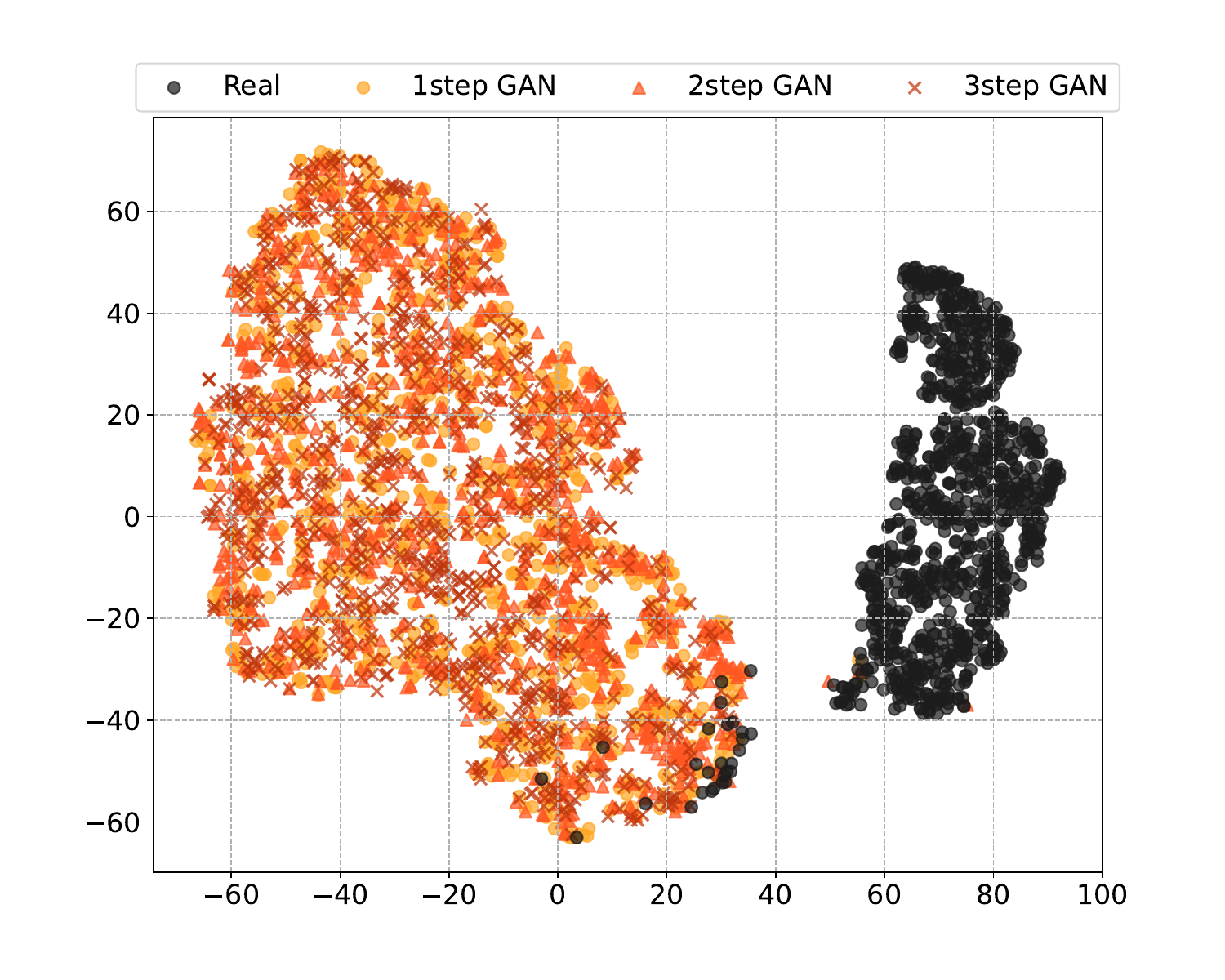}
    \caption{Train: 1-Step GAN}
    \label{fig:tsne_1stepgan}
  \end{subfigure}
  \hfill
  \begin{subfigure}[t]{0.32\linewidth}
    \includegraphics[width=\linewidth]{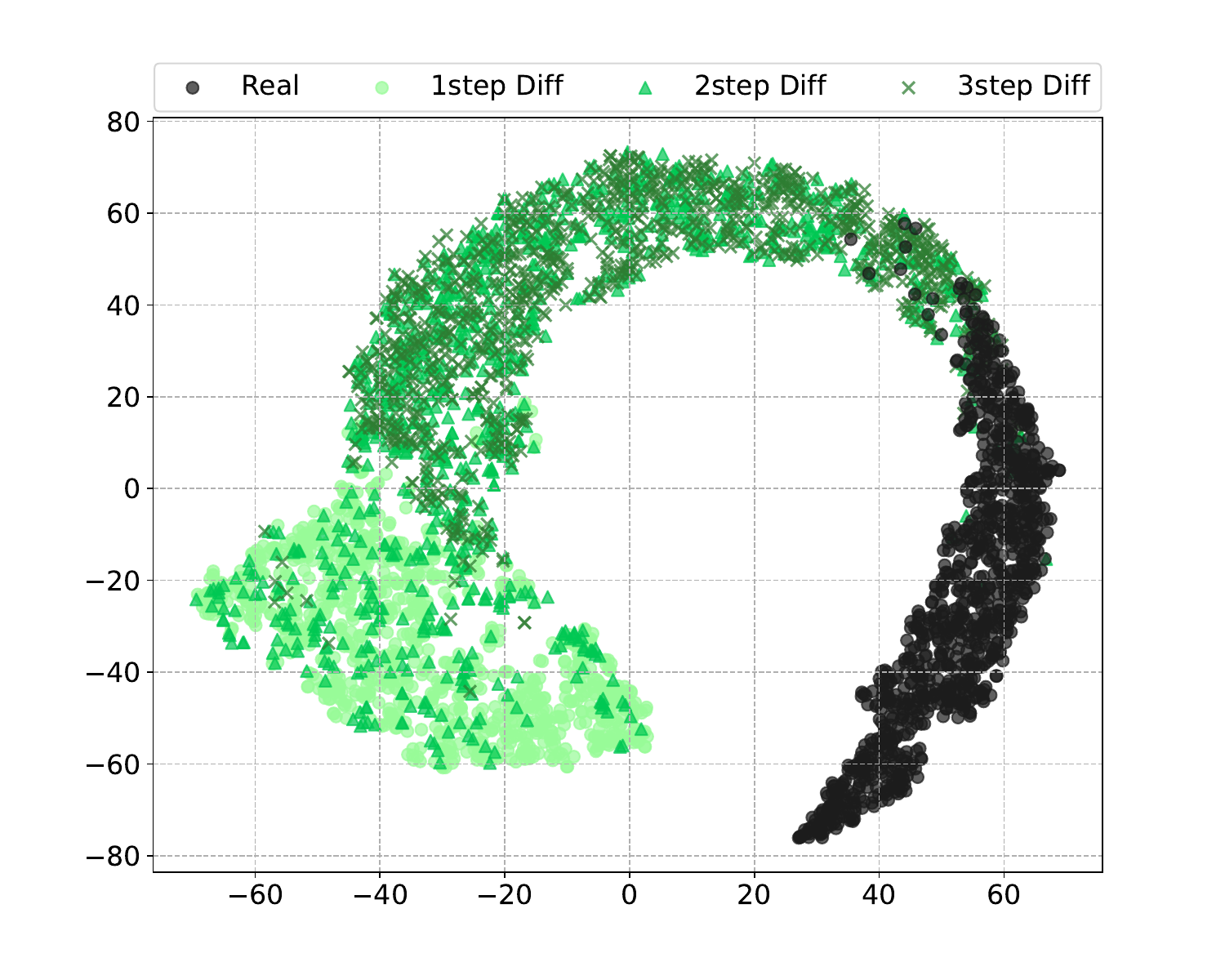}
    \caption{Train: 1-Step Diffusion}
    \label{fig:tsne_1stepdiff}
  \end{subfigure}

  \begin{subfigure}[t]{0.32\linewidth}
    \includegraphics[width=\linewidth]{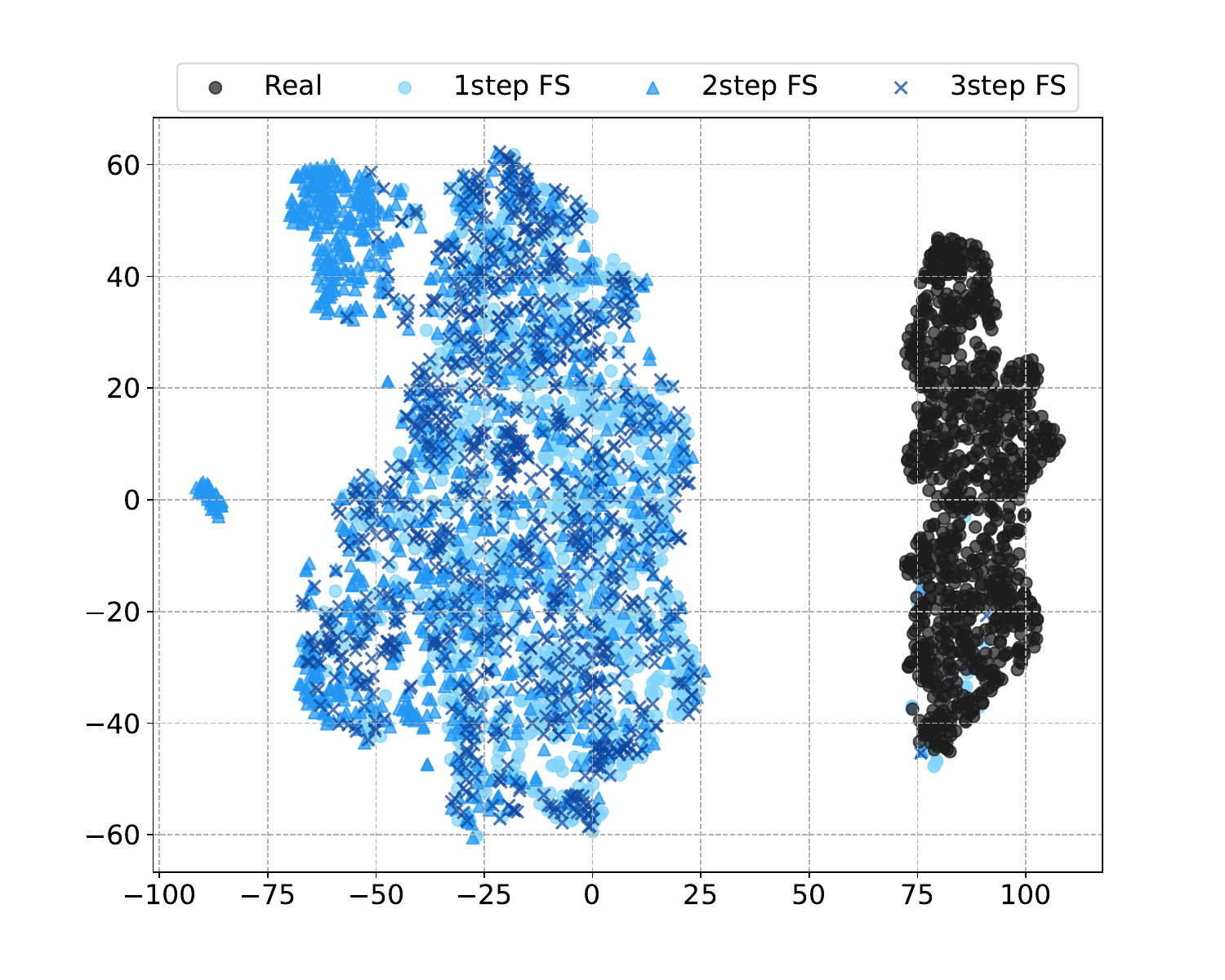}
    \caption{Train: 2-Step FaceSwap}
    \label{fig:tsne_2stepfs}
  \end{subfigure}
  \hfill
  \begin{subfigure}[t]{0.32\linewidth}
    \includegraphics[width=\linewidth]{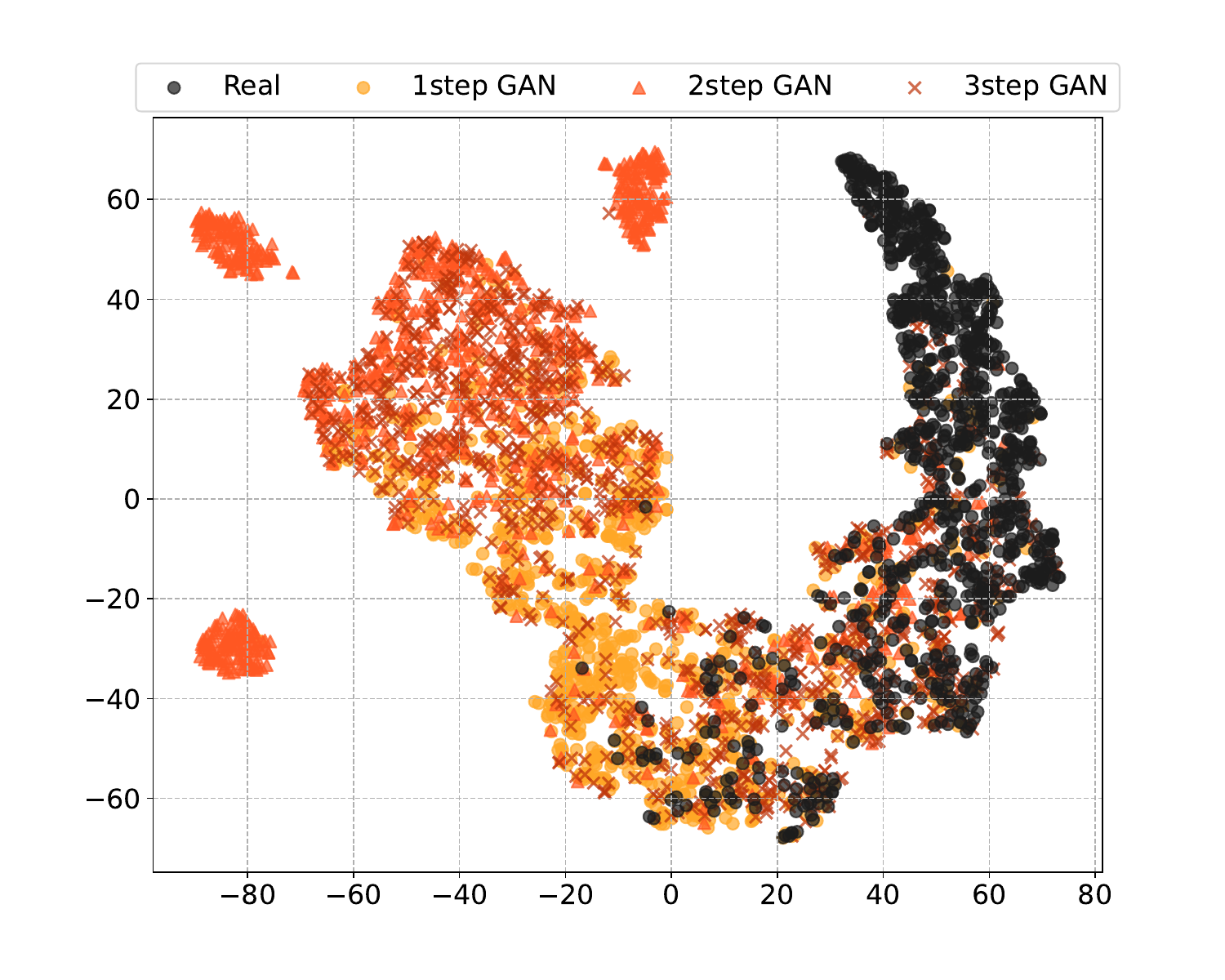}
    \caption{Train: 2-Step GAN}
    \label{fig:tsne_2stepgan}
  \end{subfigure}
  \hfill
  \begin{subfigure}[t]{0.32\linewidth}
    \includegraphics[width=\linewidth]{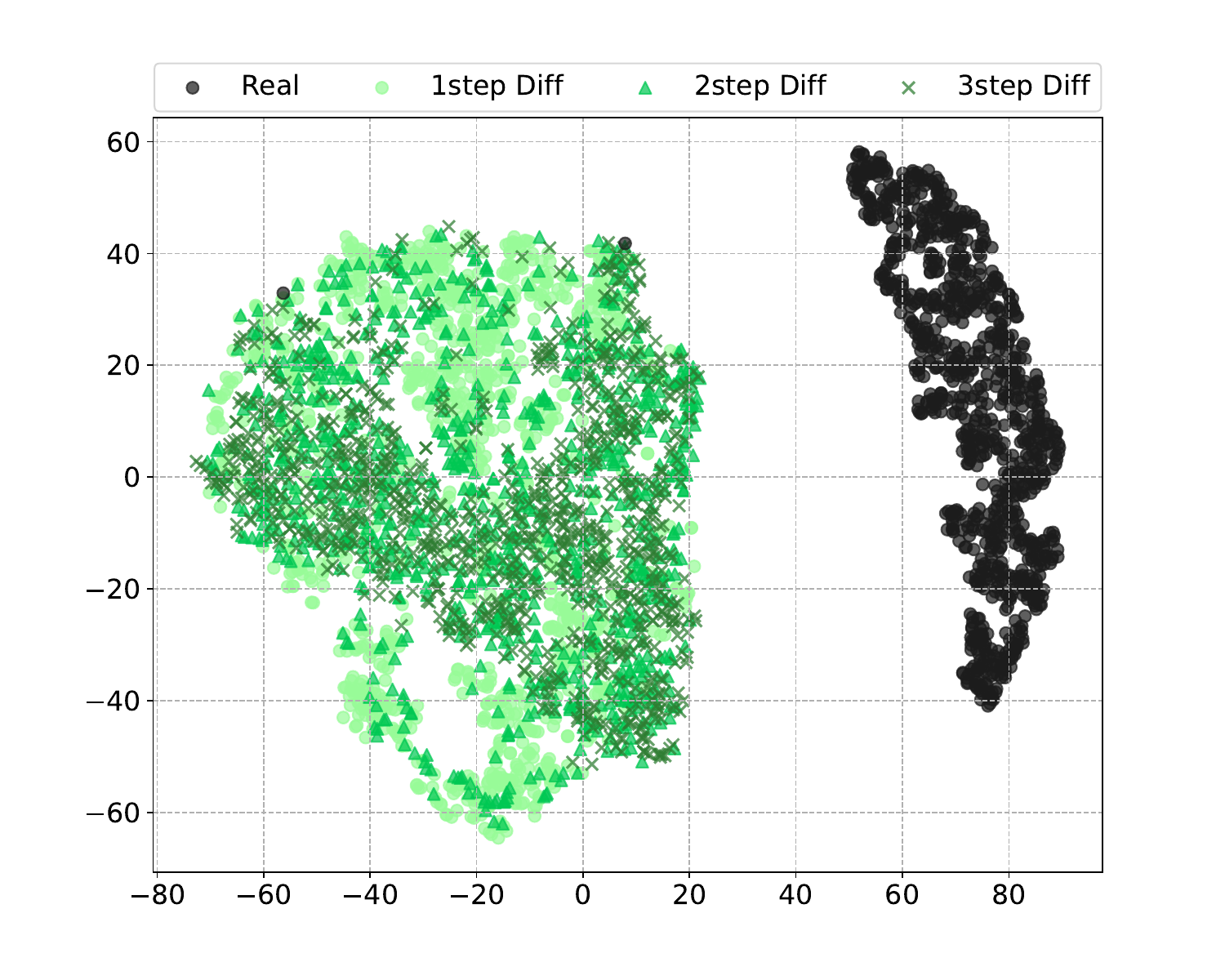}
    \caption{Train: 2-Step Diffusion}
    \label{fig:tsne_2stepdiff}
  \end{subfigure}

  \caption{
t-SNE visualizations of penultimate-layer features for real and manipulated images across different manipulation types and training settings. Each column corresponds to a manipulation type: \textcolor{blue}{FaceSwap} (blue), \textcolor{red}{GAN} (orange), and \textcolor{green!60!black}{Diffusion} (green). 
Top row (a–c) shows features extracted from models trained on 1-Step manipulated data, while the bottom row (d–f) shows results from models trained on 2-Step manipulated data. Black points represent real images. Each plot illustrates how well different manipulation depths (1-, 2-, 3-Step) are clustered or separated from real samples under various training regimes.
}

  \label{fig:tsne}
\end{figure*}


\subsection{Quantitative Results}

This observation suggests a clear divide in how synthesis models treat prior content. While GANs and Diffusion models aggressively replace existing structures in both spatial and frequency domains, FaceFusion operates more conservatively, invariably preserving residual signals from earlier edits.


\subsubsection{\textbf{Detection Performance Across Manipulation Combinations}}

We evaluate the performance of three representative detectors—Xception, F3Net, and MAT—on multi-step deepfakes generated using FaceFusion, GANs, and Diffusion models. Each detector is trained separately on data where the final manipulation corresponds to one of the three types and is evaluated across all combinations of manipulation depth (1-, 2-, and 3-Step) and final generator. ~\autoref{tab:stepwise_detection1} and ~\autoref{tab:stepwise_detection2} present the AUC and F1 scores of models trained on 1-Step and 2-Step manipulated data, respectively.

\noindent\textbf{\textit{Impact of the Final Generator on Detection Accuracy}}
Detection results show that performance is strongly influenced by the final manipulation step. Matching the final-step manipulation between training and test samples consistently leads to high AUC and F1-score, regardless of prior manipulations. For example, F3Net trained on 1-Step FaceSwap achieves over 97\% F1-score on 2- and 3-Step FaceSwap fakes. Similar trends are observed for detectors trained on samples where GANs or Diffusion are applied as the final manipulation step, highlighting the dominant influence of final-stage artifacts.

\noindent\textbf{\textit{Training Depth vs. Generalization Gap}}
While the final step governs detectability, the training depth significantly affects generalization, with distinct patterns across manipulation types.

\begin{itemize}
    \item \textbf{FaceSwap:} Detectors trained on either 1- or 2-Step data generalize well across depths, indicating consistent artifact preservation.
    
    \item \textbf{GAN:} 1-Step training yields the best generalization, whereas models trained on 2-Step GAN often overfit and perform poorly when evaluated on 1-Step data, with F3Net's F1-score dropping from 94.87\% to 39.64\%.
    
    \item \textbf{Diffusion:} Shows the opposite trend. 1-Step training results in poor generalization, while 2-Step training enables robust performance across all depths.
\end{itemize}
These patterns confirm that optimal training depth is method-specific.

\noindent\textbf{\textit{Implications for Robust Detection Design}}
Our findings suggest that detectors often overfit to shallow, final-step-specific cues and fail under composition shifts. For robust generalization, models should be trained with diverse manipulation chains that span both generator types and manipulation depths. Addressing this limitation requires reasoning over compositional histories rather than relying on static surface-level features—a direction explored further in our feature and frequency analysis.

\subsubsection{\textbf{t-SNE feature visualization}}

\autoref{fig:tsne} presents a qualitative analysis of feature embeddings extracted from the Xception model under various training conditions, illustrating how well manipulation types and depths are separated in feature space. Feature embeddings are visualized for six training setups, where the model is trained on either 1-Step or 2-Step data from the FaceSwap, GAN, or Diffusion categories, and tested on 1-, 2-, and 3-Step manipulations within the same category. The resulting clustering patterns provide insight into the model’s ability to generalize, closely aligning with the trends observed in detection performance.

\noindent\textbf{\textit{FaceSwap: Consistent Generalization Across Manipulation Depths.}}
In both training settings, whether the model is trained on 1-Step or 2-Step FS data, the Xception model effectively separates all manipulation depths (1-Step, 2-Step, and 3-Step FS) from the real class.  As shown in ~\autoref{fig:tsne}(a) and ~\autoref{fig:tsne}(d), manipulated samples form a unified cluster distinct from real samples, demonstrating that the model captures consistent forensic features specific to FaceFusion. This visual separability reinforces our detection results, where FS-trained detectors generalized well across all depths when the final step was FaceSwap (FaceFusion).

\noindent\textbf{\textit{GAN: Strong Cross-Depth Generalization from 1-Step Training.}}
~\autoref{fig:tsne}(b), where the model is trained on 1-Step GAN data, shows clear separation of all manipulated samples from the real class. The 1-Step, 2-Step, and 3-Step GAN samples form a coherent cluster, suggesting strong generalization across depths. However, in ~\autoref{fig:tsne}(e), training on 2-Step GAN data leads to overfitting: only 2-Step GAN samples remain clearly separated, while 1-Step and 3-Step GAN samples overlap significantly with real samples. This aligns with detection performance trends, where detectors trained on 2-Step GANs showed poor generalization to other depths.

\noindent\textbf{\textit{Diffusion: Enhanced Separability Through 2-Step Training.}}
~\autoref{fig:tsne}(c) indicates that training on 1-Step Diffusion data results in good separation for 1-Step samples, but 2-Step and 3-Step manipulated samples partially overlap with real data, forming bridge-like transitions. In contrast, ~\autoref{fig:tsne}(f) shows that training on 2-Step Diffusion data enables better separation of all three manipulation depths from real samples. The consistent separation in t-SNE space mirrors our performance results, where 2-Step Diffusion training led to strong generalization across manipulation depths.

Overall, the t-SNE visualizations support the conclusion that effective training depth varies by generator type: FaceSwap-trained models are robust across all depths, GAN-based models perform best when trained on 1-Step data, and Diffusion-based detectors benefit from 2-Step training. These findings emphasize the importance of tailoring training strategies to manipulation-specific characteristics.

\begin{figure}[t!]
  \centering
  
  \begin{subfigure}[t]{0.95\linewidth}
    \includegraphics[width=\linewidth]{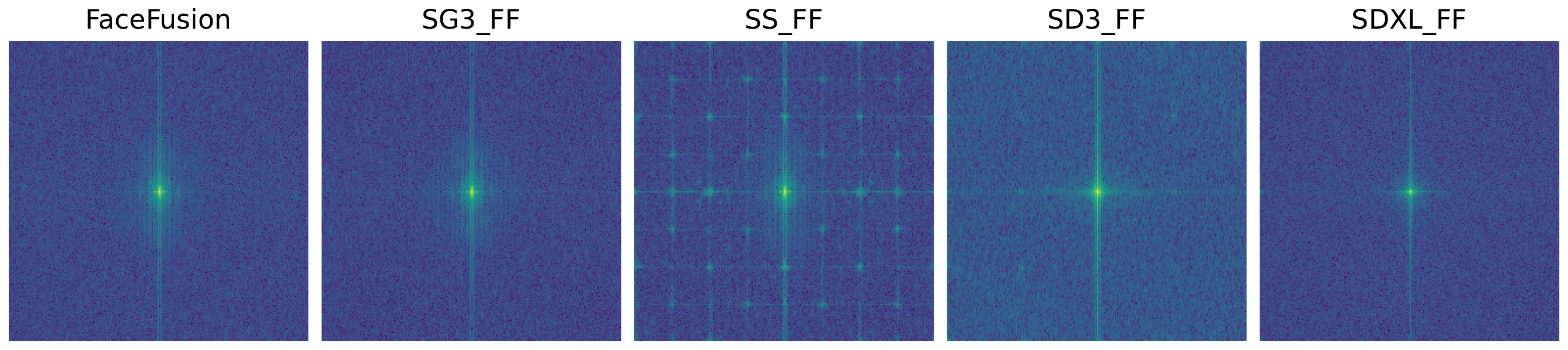}
    \caption{Final: FaceFusion}
    \label{fig:freq_2step_ff}
  \end{subfigure}


  \begin{subfigure}[t]{0.95\linewidth}
    \includegraphics[width=\linewidth]{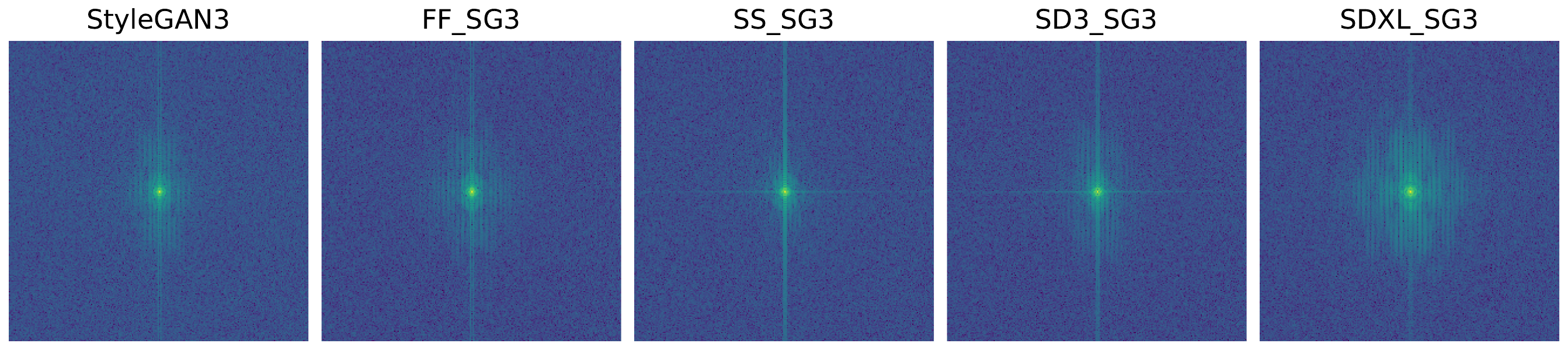}
    \caption{Final: StyleGAN3}
    \label{fig:freq_single}
  \end{subfigure}


  \begin{subfigure}[t]{0.95\linewidth}
    \includegraphics[width=\linewidth]{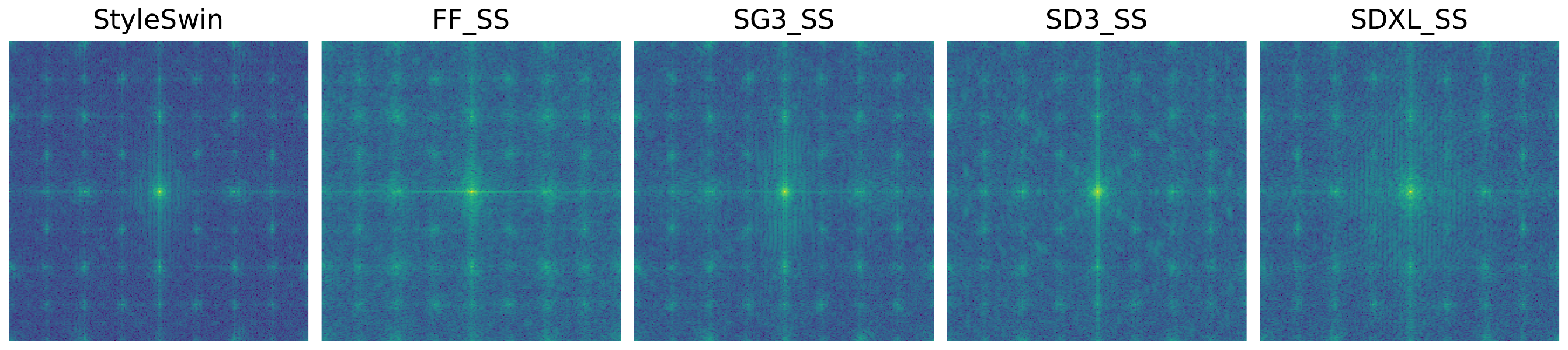}
    \caption{Final: StyleSwin}
    \label{fig:freq_2step_sg3}
  \end{subfigure}


  \begin{subfigure}[t]{0.95\linewidth}
    \includegraphics[width=\linewidth]{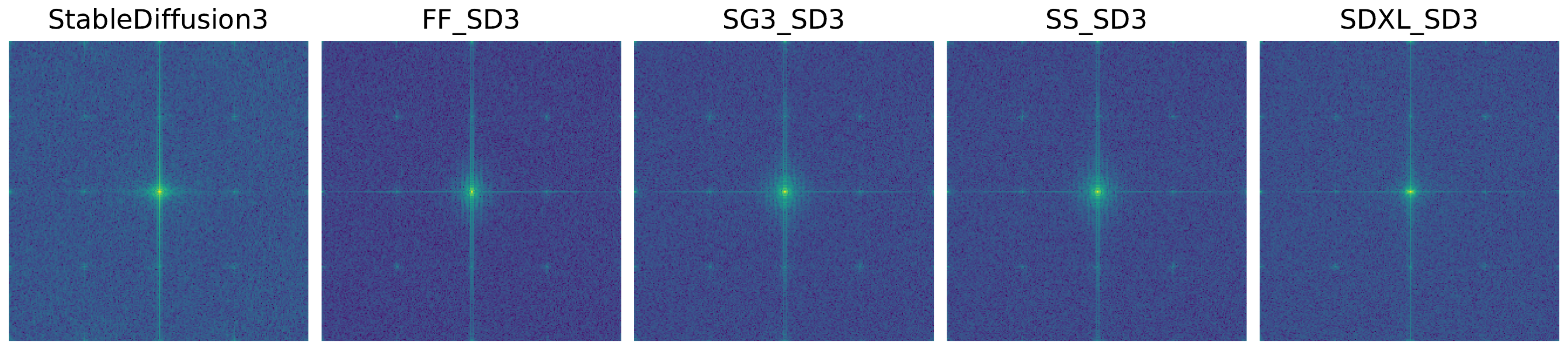}
    \caption{Final: Stable Diffusion3}
    \label{fig:freq_2step_sd3}
  \end{subfigure}


  \begin{subfigure}[t]{0.95\linewidth}
    \includegraphics[width=\linewidth]{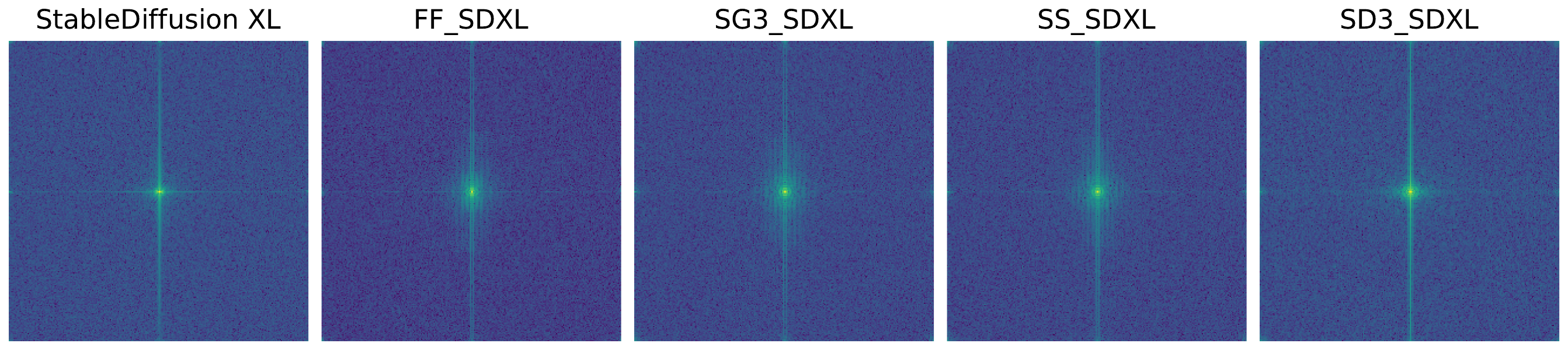}
    \caption{Final: Stable Diffusion XL}
    \label{fig:freq_2step_sdxl}
  \end{subfigure}

  \caption{
Averaged FFT spectra when each manipulation method is applied as the final step. In each subfigure, the leftmost image shows the frequency response of a 1-Step manipulation, while the four images to the right represent 2-Step manipulations that share the same final generator but differ in the preceding generation stage.
}
  \label{fig:freq_all}
\end{figure}

\subsubsection{\textbf{Spectral Overwriting in Frequency Domain}}
We analyze the spectral characteristics of manipulated images using FFT to determine whether final manipulations overwrite earlier frequency patterns. Figure~\ref{fig:freq_all} presents averaged frequency spectra for each manipulation method when applied as the final step. Each row begins with a 1-Step sample, followed by four 2-Step compositions using the same method in the final stage.

\noindent\textbf{\textit{GANs and Diffusion: Frequency-Level Overwriting.}}  
In both GAN and Diffusion settings, the final manipulation dominates the frequency spectrum. The spectral signatures of earlier manipulations are largely suppressed, suggesting that these generators aggressively replace structural information.

\noindent\textbf{\textit{FaceSwap: Residual Frequency Preservation.}}  
In contrast, FaceFusion retains frequency patterns from previous manipulations even when applied last. This behavior is evident in 2-Step samples where FaceFusion follows GAN- or Diffusion methods. The result indicates a conservative generation trend that preserves prior content.

These spectral patterns reveal consistent trends in detection performance and feature clustering, both shaped by the type of manipulation applied at the final step, highlighting the dominance of final-stage edits. 


\begin{table}[]
\begin{tabular}{l|cc|l}
\hline
\multirow{2}{*}{\textbf{\textsc{Final}}} & \multicolumn{2}{c|}{\textbf{\textsc{Inter-Step MI}}} & \multirow{2}{*}{\textbf{MI Drop}} \\ \cline{2-3}& \(\textbf{1-Step} \leftrightarrow \textbf{2-Step}\) & \(\textbf{1-Step} \leftrightarrow \textbf{3-Step}\) \\ \hline
FF         & 0.0143        & 0.0121        & 0.0022 \\
SG3          & 0.0130        & 0.0125        & 0.0006 \\
SS          & 0.0130        & 0.0127        & 0.0004 \\
SD3  & 0.0133        & 0.0118        & 0.0015 \\
SDXL & 0.0136        & 0.0134        & 0.0003 \\ \hline
\end{tabular}
\caption{Mutual information between 1-step and deeper manipulation stages. 
Each row corresponds to the method applied at the \textit{final step} of a multi-step manipulation pipeline.}
\label{tab:interstep_mi}
\end{table}

\subsubsection{\textbf{Mutual Information Analysis}}

We investigate how multi-step manipulations affect the retention of early-stage information by computing the mutual information (MI) between features at the 1-Step stage and those at subsequent stages (2-Step and 3-Step). Table~\ref{tab:interstep_mi} shows a consistent decrease in MI from 1-Step–2-Step to 1-Step–3-Step across all manipulation types, indicating a progressive loss of information as manipulations deepen. This erosion of early-stage information supports the broader finding that the final manipulation step plays a dominant role in shaping both detectability and representation, as deeper manipulations progressively overwrite or obscure initial traces. These results highlight the need for detectors to explicitly account for manipulation depth and sequence to remain effective against complex, multi-step forgeries.



\begin{table}[]
\centering
\resizebox{\columnwidth}{!}{
\begin{tabular}{llllllllll}
\hline
\multicolumn{10}{c}{\cellcolor[HTML]{D9D9D9}\textbf{\textsc{raw}}} \\ \hline
\multicolumn{1}{l|}{} & \multicolumn{3}{c|}{\textbf{1-Step}} & \multicolumn{3}{c|}{\textbf{2-Step}} & \multicolumn{3}{c}{\textbf{3-Step}} \\
\multicolumn{1}{l|}{\multirow{-2}{*}{\textbf{\textsc{Detectors}}}} & \textit{FS} & \textit{GAN} & \multicolumn{1}{l|}{\textit{Diff}} & \textit{FS} & \textit{GAN} & \multicolumn{1}{l|}{\textit{Diff}} & \textit{FS} & \textit{GAN} & \textit{Diff} \\ \hline
\multicolumn{1}{l|}{Xception} & 98.42 & 98.41 & \multicolumn{1}{l|}{98.76} & 98.72 & 97.8 & \multicolumn{1}{l|}{98.76} & 98.76 & 98.54 & 98.76 \\
\multicolumn{1}{l|}{F3Net} & 93.76 & 81.55 & \multicolumn{1}{l|}{94.02} & 94.35 & 79.17 & \multicolumn{1}{l|}{94.83} & 94.89 & 79.88 & 94.25 \\
\multicolumn{1}{l|}{MAT} & \textbf{98.82} & \textbf{99.88} & \multicolumn{1}{l|}{\textbf{99.86}} & \textbf{99.87} & \textbf{99.88} & \multicolumn{1}{l|}{\textbf{99.88}} & \textbf{99.88} & \textbf{99.88} & \textbf{99.88} \\ \hline
\multicolumn{10}{c}{\cellcolor[HTML]{D9D9D9}\textbf{\textsc{jpeg 75}}} \\ \hline
\multicolumn{1}{l|}{} & \multicolumn{3}{c|}{\textbf{1-Step}} & \multicolumn{3}{c|}{\textbf{2-Step}} & \multicolumn{3}{c}{\textbf{3-Step}} \\
\multicolumn{1}{l|}{\multirow{-2}{*}{\textbf{\textsc{Detectors}}}} & \textit{FS} & \textit{GAN} & \multicolumn{1}{l|}{\textit{Diff}} & \textit{FS} & \textit{GAN} & \multicolumn{1}{l|}{\textit{Diff}} & \textit{FS} & \textit{GAN} & \textit{Diff} \\ \hline
\multicolumn{1}{l|}{Xception} & \textbf{98.12} & \textbf{92.18} & \multicolumn{1}{l|}{\textbf{99.43}} & \textbf{98.74} & \textbf{91.88} & \multicolumn{1}{l|}{\textbf{99.39}} & \textbf{99.16} & \textbf{92.07} & \textbf{99.36} \\
\multicolumn{1}{l|}{F3Net} & 93.10 & 71.20 & \multicolumn{1}{l|}{82.23} & 88.06 & 72.09 & \multicolumn{1}{l|}{93.40} & 91.59 & 70.93 & 90.30 \\
\multicolumn{1}{l|}{MAT} & 56.30 & 56.39 & \multicolumn{1}{l|}{56.42} & 56.42 & 56.42 & \multicolumn{1}{l|}{56.42} & 56.42 & 56.42 & 56.42 \\ \hline
\multicolumn{10}{c}{\cellcolor[HTML]{D9D9D9}\textbf{\textsc{jpeg 75x2}}} \\ \hline
\multicolumn{1}{l|}{} & \multicolumn{3}{c|}{\textbf{1-Step}} & \multicolumn{3}{c|}{\textbf{2-Step}} & \multicolumn{3}{c}{\textbf{3-Step}} \\
\multicolumn{1}{l|}{\multirow{-2}{*}{\textbf{\textsc{Detectors}}}} & \textit{FS} & \textit{GAN} & \multicolumn{1}{l|}{\textit{Diff}} & \textit{FS} & \textit{GAN} & \multicolumn{1}{l|}{\textit{Diff}} & \textit{FS} & \textit{GAN} & \textit{Diff} \\ \hline
\multicolumn{1}{l|}{Xception} & \textbf{98.10} & \textbf{92.17} & \multicolumn{1}{l|}{\textbf{99.43}} & \textbf{98.74} & \textbf{91.85} & \multicolumn{1}{l|}{\textbf{99.38}} & \textbf{99.16} & \textbf{92.07} & \textbf{99.36} \\
\multicolumn{1}{l|}{F3Net} & 93.12 & 71.22 & \multicolumn{1}{l|}{82.26} & 88.08 & 72.12 & \multicolumn{1}{l|}{93.44} & 91.62 & 70.91 & 90.33 \\
\multicolumn{1}{l|}{MAT} & 56.28 & 56.33 & \multicolumn{1}{l|}{56.36} & 56.36 & 56.36 & \multicolumn{1}{l|}{56.36} & 56.36 & 56.36 & 56.36 \\ \hline
\end{tabular}
}
\caption{Detection accuracy (\%) of models trained on the best combination (1-Step FS, 1-Step GAN, 2-Step Diff) and evaluated on test data under three compression settings: uncompressed (raw), JPEG 75, and JPEG 75 applied twice (JPEG 75×2). Results are grouped by manipulation depth (1-, 2-, and 3- step) and final manipulation type (FS, GAN, Diff).}

\label{tab:quality_detection}
\end{table}

\subsubsection{\textbf{Impact of JPEG Compression on Best-Trained Detectors}}
~\autoref{tab:quality_detection} presents detection accuracy results for models trained on the best-performing combination of training data, which includes 1-Step FaceSwap (FS), 1-Step GAN, and 2-Step Diffusion (Diff), and tested on 1-, 2-, and 3-Step manipulated images across three quality settings: raw, JPEG 75, and JPEG 75 applied twice (JPEG 75×2). Under the raw setting, both Xception and MAT models demonstrate high robustness, achieving over 97\% accuracy across all manipulation types and depths. However, with JPEG 75 compression, a notable drop is observed for MAT, whose performance degrades sharply to the 56\% range, indicating significant sensitivity to compression artifacts. In contrast, Xception remains the best performer under JPEG 75, showing only a modest average accuracy drop of 1.84\% compared to the raw setting. Interestingly, when the compression is applied twice (JPEG 75×2), the performance of all detectors remains largely stable, with only a negligible average accuracy drop of approximately 0.01 compared to JPEG 75. This suggests repeated JPEG compression does not introduce additional degradation beyond the first compression step, reinforcing the relative resilience of certain models including Xception against moderate-to-strong compression conditions.

\subsection{Qualitative Results}

\begin{figure}[] 
  \centering
  \includegraphics[trim={0pt 2pt 2pt 0pt},clip,width=1\linewidth]{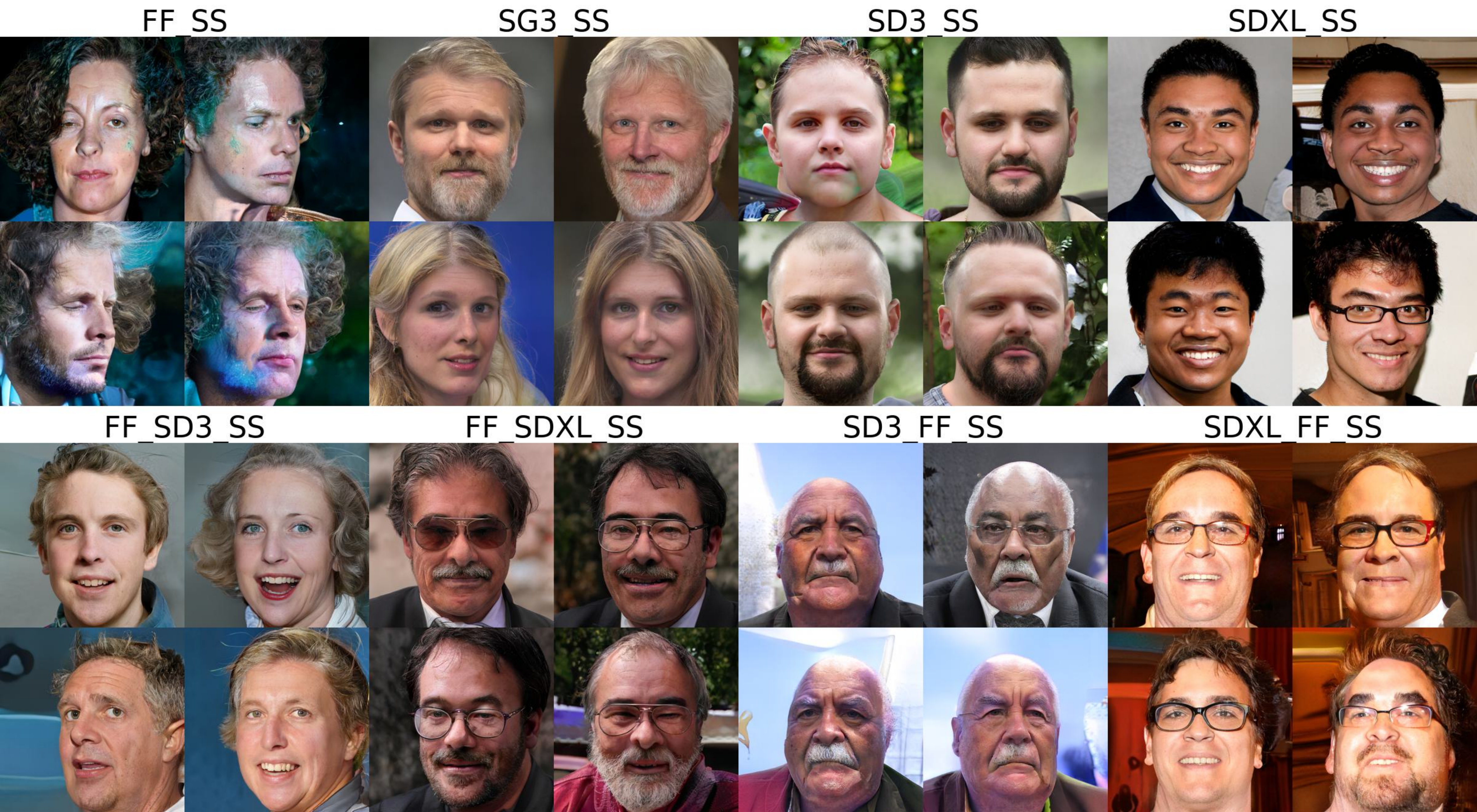}
  \caption{Using StyleSwin as the final step in multi-stage manipulations leads to identity collapse, where outputs exhibit similar facial features regardless of the input source.} 
  \label{fig:LastStep_SS}
\end{figure}

We analyzed multi-step manipulation chains to assess whether specific generators introduce consistent visual biases. While most combinations preserved diverse identities and styles, StyleSwin showed a notable exception. When used as the final step, it consistently produced biased outputs shaped by prior manipulations. For instance, when StyleSwin followed FaceFusion, it frequently generated curly-haired male faces with dark backgrounds, while pairing it with Stable Diffusion XL often resulted in Black male faces with dark hair (see \texttt{FF\_SS} and \texttt{SDXL\_SS} in ~\autoref{fig:LastStep_SS}). In contrast, manipulation paths that did not involve StyleSwin as the final generator maintained greater visual diversity. These observations suggest that StyleSwin is particularly sensitive to earlier edits and may reduce semantic variety in the final output.

One possible explanation for this phenomenon lies in the design of the StyleSwin architecture. This generator combines window-based self-attention with style modulation, and it tends to impose strong internal priors during synthesis. The style injection mechanism and the use of hierarchical attention layers may result in visual features such as background, hairstyle, and facial attributes being driven by the model’s learned distributions rather than by the preceding manipulations. Additionally, the application of double attention layers, which are designed to enhance global consistency, may further reinforce dominant patterns and diminish residual details from earlier stages. As a result, the outputs become semantically similar regardless of the input history, which explains the identity convergence observed in our experiments.

\section{Conclusion}
Our study investigates how multi-step deepfake manipulations affect deepfake detection performance. We evaluate detectors across combinations of FS-, GAN-, and Diffusion-based generators, focusing on how manipulation depth and final generator type influence performance.
Our results show that detectability is primarily influenced by the final manipulation step. Moreover, performance differs substantially depending on whether training was conducted on single-type or multi-type manipulation data, with varying trends observed across FS-, GAN-, and Diffusion-based methods. First, FS-based manipulations are robust across depths, while GAN-based detectors generalize well only when trained on 1-Step data. Diffusion-based models require 2-Step training to effectively separate manipulation depths, as confirmed by detection metrics and t-SNE analysis.
Further, frequency analysis reveals that GAN and Diffusion models tend to overwrite earlier artifacts, whereas FaceFusion preserves prior spectral patterns. Consistently, mutual information analysis confirms a progressive loss of early-stage signals across all methods, reinforcing that the final manipulation step ultimately dominates both detectability and representation. Additionally, qualitative analysis shows that the architectural characteristics of underlying generators, such as StyleSwin, can induce identity collapse during multi-step manipulations by reinforcing internal priors and suppressing semantic diversity. Our findings highlight the need for training strategies tailored to manipulation type and depth to build detectors resilient to complex, real-world forgery pipelines.


\section*{Acknowledgments}
The authors would thank anonymous reviewers. Simon S. Woo is the corresponding author. This work was partly supported by Institute for Information \& communication Technology Planning \& evaluation (IITP) grants funded by the Korean government MSIT:
(RS-2022-II221199, RS-2022-II220688, RS-2019-II190421, RS-2023-00230337, RS-2024-00356293, RS-2024-00437849, RS-2021-II212068,  RS-2025-02304983, RS-2025-02263841, RS-2024-00436936, RS-2024-00437849).


\bibliographystyle{ACM-Reference-Format}
\balance
\bibliography{references}

\end{document}